\documentclass[fleqn,10pt]{wlscirep}
\usepackage[utf8]{inputenc}
\usepackage[T1]{fontenc}
\usepackage{comment}

\usepackage{amsmath,amssymb,amsfonts}
\usepackage{algorithmic}
\usepackage{graphicx}
\usepackage{algorithm,algorithmic}
\usepackage{hyperref}
\usepackage{multirow}
\hypersetup{hidelinks=true}
\usepackage{textcomp}
\usepackage[table,xcdraw]{xcolor}
\usepackage{comment}
\usepackage{makecell}
\usepackage{booktabs}
\usepackage{tikz}
\usepackage{wrapfig}

\usepackage{subcaption}
\usepackage{svg}
\usepackage{mwe}

\title{Mammo-CLIP Dissect: A Framework for Analysing Mammography Concepts in Vision-Language Models}
\author[1,*]{Suaiba Amina Salahuddin}
\author[2]{Teresa Dorszewski}
\author[3]{Marit Almenning Martiniussen}
\author[4]{Tone Hovda}
\author[7,8]{Antonio Portaluri}
\author[1]{Solveig Thrun}
\author[1,5]{Michael Kampffmeyer}
\author[1]{Elisabeth Wetzer}
\author[1]{Kristoffer Wickstrøm}
\author[1,5,6]{Robert Jenssen}
\affil[1]{UiT The Arctic University of Norway, Department of Physics and Technology, Tromsø, 9019, Norway}
\affil[2]{Technical University of Denmark, Department of Applied Mathematics and Computer Science, Copenhagen, 2800, Denmark}
\affil[3]{Østfold Hospital Trust, department, Fredrikstad, 1714, Norway}
\affil[4]{Vestre Viken Hospital Trust, Department of Radiology, Drammen, 3004, Norway}
\affil[5]{Norwegian Computing Center, Oslo, 0373, Norway}
\affil[6]{University of Copenhagen, Pioneer Centre for AI, Copenhagen, 1350, Denmark}
\affil[7]{Radboud University Nijmegen Medical Centre, Nijmegen, 6500 HB, Netherlands}
\affil[8]{The Netherlands Cancer Institute, Antoni van Leeuwenhoek Hospital, Amsterdam, 1066 CX, Netherlands }

\affil[*]{ssa195@uit.no}

\begin{document}
\begin{abstract}
Understanding what deep learning (DL) models learn is essential for the safe deployment of artificial intelligence (AI) in clinical settings. While previous work has primarily focused on pixel-based explainability methods, less attention has been paid to the textual concepts learned by these models, which may more closely reflect the reasoning used by clinicians. We introduce Mammo-CLIP Dissect, the first concept-based explainability framework for systematically dissecting DL vision models trained for mammography. Leveraging a mammography-specific vision–language model (Mammo-CLIP) as a “dissector,” our approach labels neurons at specified layers with human-interpretable textual concepts and quantifies their alignment to domain knowledge. 

Using Mammo-CLIP Dissect, we investigate three key questions: (1) how concept learning differs between DL vision models trained on general image datasets versus mammography-specific datasets; (2) how fine-tuning for downstream mammography tasks affects concept specialisation; and (3) which mammography-relevant concepts remain under-represented. We show that models trained on mammography data capture more clinically relevant concepts and align more closely with radiologists’ workflows than models not trained on mammography data. Fine-tuning for task-specific classification enhances the capture of certain concept categories (e.g., benign calcifications) but can lead to reduced coverage of others (e.g., density-related features), indicating a trade-off between specialisation and generalisation. At the neuron level, we demonstrate selective responsiveness of neurons to distinct mammography features such as implants and calcifications, demonstrating the framework’s ability to capture and specialise on key mammography concepts. 

Our findings show that concept-based explainability with Mammo-CLIP Dissect provides new insights into how convolutional neural networks (CNNs) capture mammography-specific knowledge. By systematically comparing models across training data and fine-tuning regimes, we reveal how domain-specific training and task-specific adaptation shape concept learning and where clinically important features remain underrepresented. This concept-level perspective moves beyond pixel-level interpretability toward an understanding more closely aligned with radiologists’ reasoning. 
Code and concept set are publicly available: \url{https://github.com/Suaiba/Mammo-CLIP-Dissect}.

\vspace{0.5cm}
\end{abstract}
\flushbottom
\maketitle
\section*{Introduction}
\begin{figure}[!ht]
    \centering
    \includegraphics[width=0.7\linewidth]{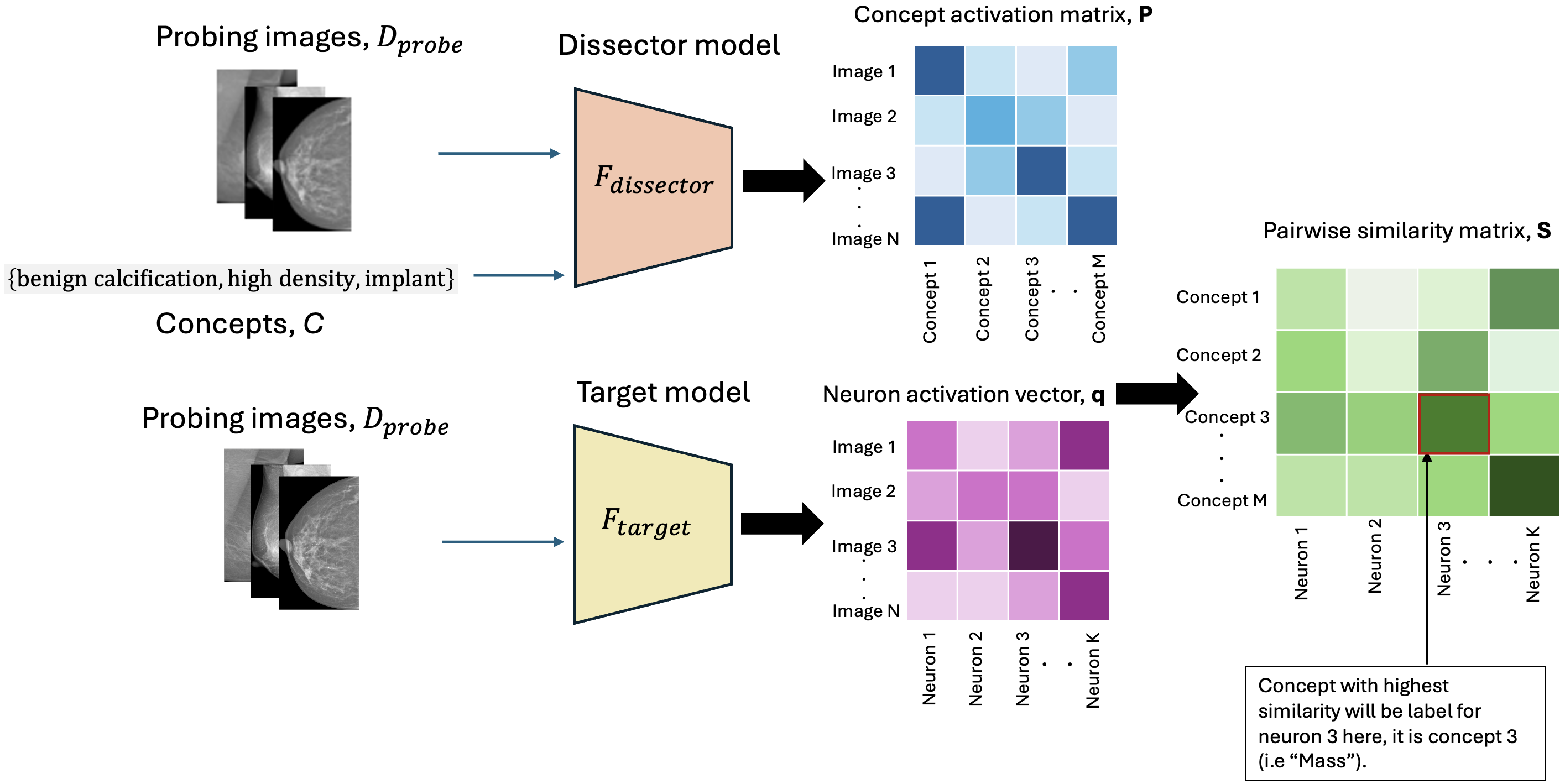}
    \caption{An overview of the Mammo-CLIP-dissect framework.}
    \label{fig:overview}
\end{figure}
Breast cancer is the most commonly diagnosed cancer and the leading cause of cancer-related deaths among women globally, with an estimated $2.3$ million diagnosed and $670,000$ deaths in 2022, and incidence continuing to rise among younger women over time \cite{Kim2025GlobalCountries,Xu_2024,Pujol2025IncreasingTime}. Early detection through X-ray-based mammography is the current gold-standard for breast cancer screening, playing a crucial role in improving treatment outcomes and overall prognosis \cite{Zielonke2020EvidenceReview}. Despite an overall reduction in breast cancer mortality due to screening, there are limitations such as misinterpretations and inadequate image quality, which can lead to failure in detecting cancer 
\cite{Hovda2021RadiologicalCancer,Houssami2017TheScreening}. Additionally, limitations in mammographic screening may lead to overdiagnosis and false positive screening results that cause anxiety and resource-intensive follow-up in asymptomatic women.

Recent advancements in artificial intelligence (AI) and deep learning (DL) have demonstrated significant potential in overcoming these issues and enhancing the accuracy and efficiency of breast cancer screening in retrospective settings \cite{dem2023,Lauritzen2022AnWorkload,Koch2024HowNorway,Wu2020DeepScreening}. Notably, the MASAI prospective study found that AI-supported readings might increase detection of clinically relevant cancers without compromising false-positive rates compared to standard readings without AI \cite{Hernstrom2025ScreeningStudy}. Such DL-based tools aim to reduce the screen–reading workload of medical practitioners and decrease false positive results by assisting in tasks such as breast cancer detection and classification, as well as the assessment of mammographic density \cite{Li2021Multi-ViewLearning,Saffari2020FullyLearning} as well as risk prediction \cite{ThrSol_Reconsidering_MICCAI2025, SunZij_VMRAMaR_MICCAI2025}.

However, despite their potential, the clinical adoption of AI technologies faces several significant challenges and concerns. One of the most pressing issues is the lack of transparency and interpretability in many AI systems \cite{Turri2024TransparencyApproaches,Longo2024ExplainableDirections}, which becomes especially problematic in safety-critical applications like healthcare and breast cancer detection. The inability to clearly understand how AI models arrive at their decisions raises concerns about trust, reliability, and accountability in clinical settings, potentially hindering their widespread acceptance and use.

The emerging field of eXplainable AI (XAI) seeks to bridge this gap by providing insights into the decision-making processes of AI systems \cite{Longo2024ExplainableDirections}. XAI identifies aspects of data that influence predictions, enabling interpretable decision-making better aligned with clinical reasoning. In mammography, the most widely adopted XAI methods—heatmaps, saliency maps, and attention mechanisms—highlight regions of interest in mammograms  \cite{Shifa2025AScreening,Ghosh2024Mammo-CLIP:Mammography,Dahl2023Two-stageDetection,Al-Tam2024MultimodalImages}. While these approaches offer useful pixel-level cues, they remain limited because they lack concept-level explanations essential for radiologists’ diagnostic reasoning. As a result, they often fail to capture contextual and relational information, such as the shape, margin, or distribution of calcifications. Furthermore, these methods can struggle to highlight global mammographic features, like high breast density, which can obscure underlying findings, since such characteristics are defined across the broader tissue distribution rather than localised pixels. Consequently, pixel-level XAI does not fully align with radiologists’ workflows, which rely on higher-level, domain-specific concepts such as masses, calcifications, and breast density.

This gap is significant because radiologists typically describe their findings in terms of semantic concepts, and the Breast Imaging Reporting and Data System (BI-RADS) \cite{Berg2023BreastSystem} is widely used for the interpretation of screening mammograms and in clinical text reports. BI-RADS categories encapsulate concepts such as mass, calcification, architectural distortion, and breast density, which are central to diagnostic decisions and communication of risk. Yet most current XAI approaches in mammography fail to bridge the divide between low-level visual features and these high-level, clinically meaningful concepts.

DL models for computer vision, such as convolutional neural networks (CNNs) \cite{Krizhevsky2012ImageNetNetworks, Simonyan2014VeryRecognition} and vision transformers (ViTs) \cite{DosovitskiyANSCALE}, are the foundation for AI-based mammography analysis. Understanding how these DL models learn mammography-specific concepts, leveraged from the radiology reports and BI-RADS descriptions, is a key step toward examining the models' reasoning, in a manner aligned with the radiologists' workflow and interpretative processes, and increases the trust in AI systems thanks to its interpretability.


Recently, vision-language models (VLMs), such as Contrastive Language-Image Pre-training (CLIP) \cite{Radford2021LearningSupervision}, have gained attention as a potential solution to the lack of concept-level XAI for mammography. By jointly aligning textual and visual information, VLMs enable models to learn associations between images and domain-specific concepts, which can later be leveraged to facilitate interpretability (e.g., through CLIP Dissect\cite{Oikarinen2022CLIP-Dissect:Networks}). In the context of mammography, Mammo-CLIP \cite{Ghosh2024Mammo-CLIP:Mammography} represents a significant advancement as the first VLM specifically designed for this domain. While classifier models based on the Mammo-CLIP image encoder have demonstrated strong performance in breast cancer detection and localisation tasks, their interpretability—particularly in terms of the mammographic concepts it learns—remains underexplored.

To address these limitations, we propose Mammo-CLIP Dissect, a novel framework
for dissecting DL-based vision models when processing mammograms. An overview of Mammo-CLIP Dissect is presented in Figure \ref{fig:overview}. In our work, we focus on vision models in the form of CNNs since such models are most common in DL-based analysis of mammography, but our study can be extended to vision models in the form of ViTs. We aim to better understand and distinguish what concepts are learnt by different CNNs, as well as analyse how the learned concepts relate to radiologists' workflows. Our work builds on two key prior studies. First,
the CLIP-Dissect framework \cite{Oikarinen2022CLIP-Dissect:Networks}, which labels neurons by concepts encoded by them within CNNs and ViTs. Second, a more recent study \cite{Dorszewski2025FromTransformers}, which leveraged CLIP-Dissect to perform a layer-wise
analysis of concepts in ViTs. This study revealed that ViTs as well as CNNs encode increasingly complex concepts across
layers, with early layers focusing on basic features (e.g., colours and textures) and later layers capturing more specific and
diverse categories (e.g., objects and animals).
Our approach builds on the strengths of these methodologies while addressing the unique challenges within the mammography
domain. By introducing a clinically informed concept set derived from radiology reports (in collaboration with radiologists) and performing analysis at the neuron-level, Mammo-CLIP Dissect provides new insights into how mammography-specific semantic concepts are learned and represented.

Our work is structured around three key research questions:
\begin{itemize}
 \item Do CNNs trained on mammography data learn more mammography-specific concepts compared to models trained on non-mammography data?

 \item How does fine-tuning for mammography-relevant tasks, such as density classification, affect the learning of mammography-specific concepts?

 \item  Which key mammography concepts are learned and which are not picked up by CNNs?
\end{itemize}
These research questions are key to understanding how DL vision models capture and represent mammography-specific knowledge. By addressing them, we can evaluate the impact of training and fine-tuning on concept learning, identify gaps between model reasoning and radiologists’ workflows, and inform the development of more interpretable and clinically aligned AI systems.

Our main contributions are as follows:
\begin{itemize}
\item Framework Development: We introduce the first framework, Mammo-CLIP dissect, for dissection of DL vision models at specified layers leveraging mammography-specific vision-language models (VLMs). By applying neuron labelling, Mammo-CLIP Dissect reveals the concepts encoded by individual neurons, offering insights into how DL vision models represent mammographic features. Specifically, we employ Mammo-CLIP as the dissector to analyse different DL vision models. This framework extends the explainability method, CLIP-dissect, used in general VLMs for natural images to the domain of mammography-specific VLMs. Our goal is to analyse the concepts learned by these mammography-specific VLMs and explore their relevance to clinical interpretations, thereby bridging the gap between DL model understanding and domain-specific medical insights.

\item Clinically Informed Concept Set: We present a new concept set derived from mammography radiology reports and BI-RADS descriptions. This set has been categorised into clinically meaningful groups to facilitate analysis in collaboration with radiologists.

\item Concept-Level Explainability for DL vision models in mammography: We provide the first concept-level explanations for the vision branch of Mammo-CLIP by leveraging neuron labelling. Our analysis explores how mammography-specific concepts are learned across DL vision models trained on mammography data and DL vision models not trained on mammography data, offering insights into the feature extraction process and its alignment with clinical workflows. We also evaluate how fine-tuning affects the learning of task-relevant concepts, highlighting shifts in representations and potential trade-offs.
\end{itemize}

\section*{Related works}
\subsection*{XAI in Mammography}
XAI methods have been applied in mammography to improve the interpretability of DL-based systems. Most current approaches rely on pixel-level explanations—such as Grad-CAM \cite{Selvaraju2017Grad-CAM:Localization}, saliency maps, and attention mechanisms—that highlight regions of interest on mammograms \cite{Shifa2025AScreening}. For example, Dahl et al. \cite{Dahl2023Two-stageDetection} used a two-stage ResNet101 \cite{he2016deep} model on the BreastScreen Norway dataset with Layered Grad-CAM to visualise regions associated with malignancies, while Raghavan et al. \cite{Raghavan2024AttentionDetection} applied attention-guided Grad-CAM to breast images using DenseNet \cite{huang2017densely}, VGG \cite{Simonyan2014VeryRecognition}, and EfficientNet alongside interactive question-answering for radiologists. Similarly, Farrag et al. \cite{Farrag2023AnSegmentation} used Grad-CAM and Occlusion Sensitivity with DeeplabV3+ \cite{chen2018encoder} on INbreast \cite{Moreira2012INbreast:Database} to validate tumour localisation, and Pertuz et al. \cite{Pertuz2023SaliencyIntelligence} compared Grad-CAM, Grad-CAM++, and Eigen-CAM to board-certified radiologists’ annotations.

Pixel-based XAI methods, such as Grad-CAM and saliency maps, lack concept-level explanations, failing to map DL model decisions to domain-specific concepts \cite{Adebayo2018SanityMaps,Cerekci2024QuantitativeAnalysis}. This limitation undermines human interpretation, as these explanations often misalign with clinical expertise and overlook critical findings, such as small calcifications essential for breast cancer diagnosis. Additionally, these methods frequently highlight irrelevant areas, like dense tissue or background structures, rather than clinically significant regions, exacerbating challenges in cases with dense breast tissue where lesions are already obscured. They also neglect contextual and relational information, such as the shape, margin, and distribution of calcifications, which are vital for malignancy assessment. To address these gaps, concept-based XAI approaches that integrate textual explanations using diagnostically relevant terms could improve interpretability and trust by aligning AI outputs with clinical reasoning.
\subsection*{CLIP}
CLIP \cite{Radford2021LearningSupervision} is one if the most widely used VLMs that jointly trains an image encoder and a text encoder on large-scale natural image-text pairs to align similar image and text
representations using contrastive learning. While CLIP is designed to generalise across diverse downstream tasks after pretraining on large-scale datasets, it is not specifically trained on medical images and has likely only encountered such data incidentally. Moreover, it has never been exposed to domain-specific contexts like mammography reports. This lack of domain-specific training raises concerns about its applicability in mammography, where minute details are critical for accurate clinical interpretation.
\subsection*{Mammo-CLIP}
Mammo-CLIP\cite{Ghosh2024Mammo-CLIP:Mammography} is a domain-specific VLM, designed to process high-resolution mammography images and corresponding radiology reports. Unlike standard CLIP, Mammo-CLIP incorporates an EfficientNet-B5 image encoder and a BioClinicalBERT text encoder, both of which were trained on large-scale, domain-specific mammography datasets. This adaptation enables Mammo-CLIP to capture subtle and clinically relevant features essential for mammography analysis. The implementation of Mammo-CLIP as well as the pre-trained models, is publicly available\footnote{\url{https://github.com/batmanlab/Mammo-CLIP}}, 
making it accessible for implementation and adaptation. While Mammo-CLIP has been applied to downstream tasks such as mass, calcification, and BI-RADS density classification, our study extends its utility by investigating the concepts learned at different layers of the model.
\subsection*{CLIP-dissect}
CLIP-dissect \cite{Oikarinen2022CLIP-Dissect:Networks} is an open-source and versatile framework\footnote{\url{https://github.com/Trustworthy-ML-Lab/CLIP-dissect}} which automatically labels neurons within DL models using a predefined set of
concepts. CLIP-dissect uses a probing dataset, a target model to dissect, and a VLM dissector (e.g., CLIP) to first compute concept-activation matrices based on similarities between images and concepts. This matrix acts as a look-up dictionary, which is used to assign labels to neurons based on their activations across probing images and similarity to predefined concepts. Dorszewski et. al. \cite{Dorszewski2025FromTransformers} adapted CLIP-dissect to enable dissection of a broader range of DL models, and the code for this study is publicly available \footnote{\url{https://github.com/teresa-sc/concepts_in_ViTs}}. 

While CLIP-Dissect has been applied to general VLMs, our study adapts this framework to the mammography domain by leveraging Mammo-CLIP and introducing a clinically informed, mammography-specific concept set.

\section*{Methods}
\begin{figure}[!ht]
    \centering
        \includegraphics[width=0.6\linewidth]{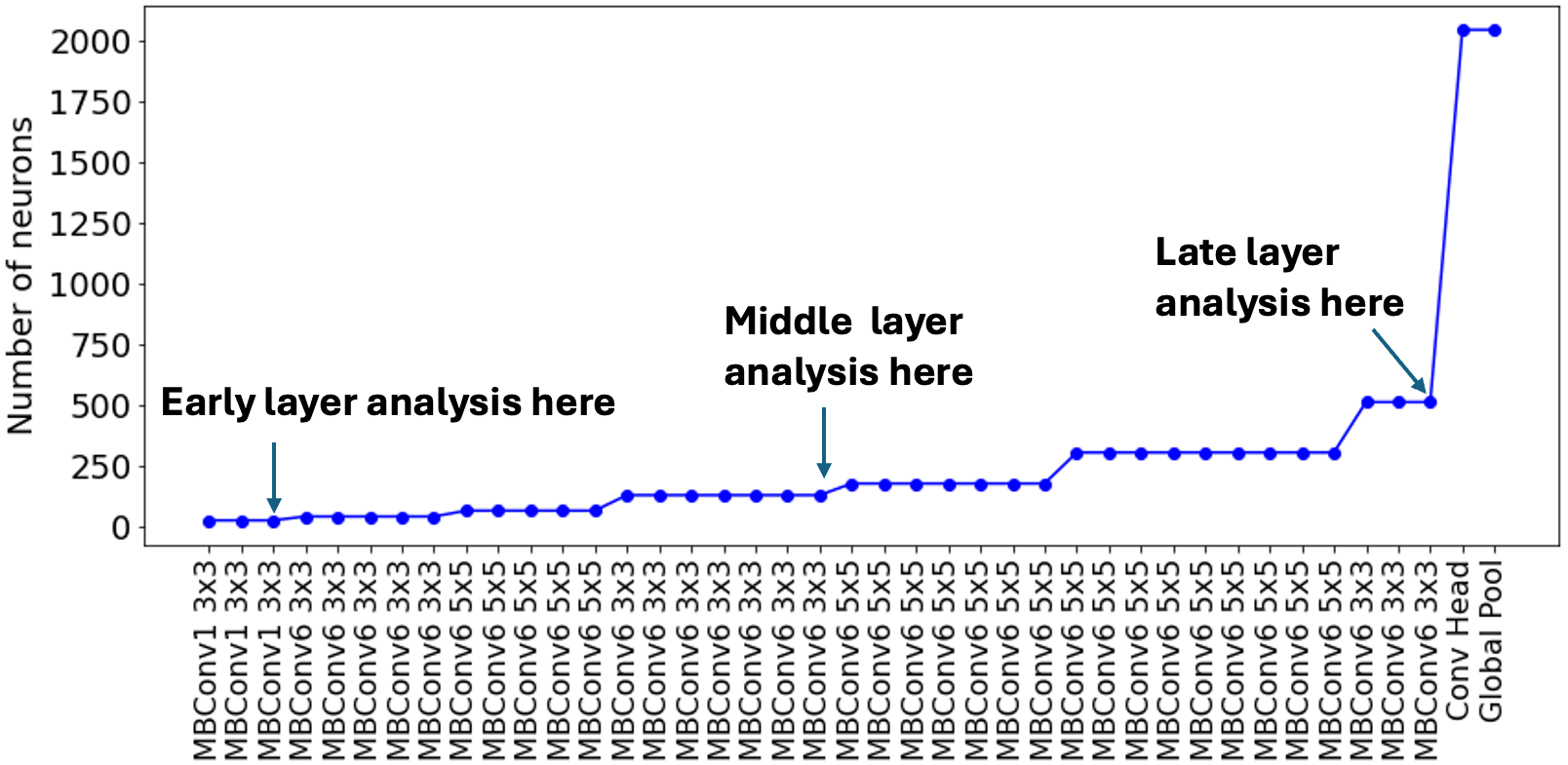}
        \caption{Structure of EfficientNet-B5 model and the grouping of layers for analysis.}
        \label{fig:effb5_standard}
\end{figure}
\begin{figure}[!ht]
    \centering
    \begin{subfigure}{0.45\textwidth} 
        \centering
        \includegraphics[width=\linewidth]{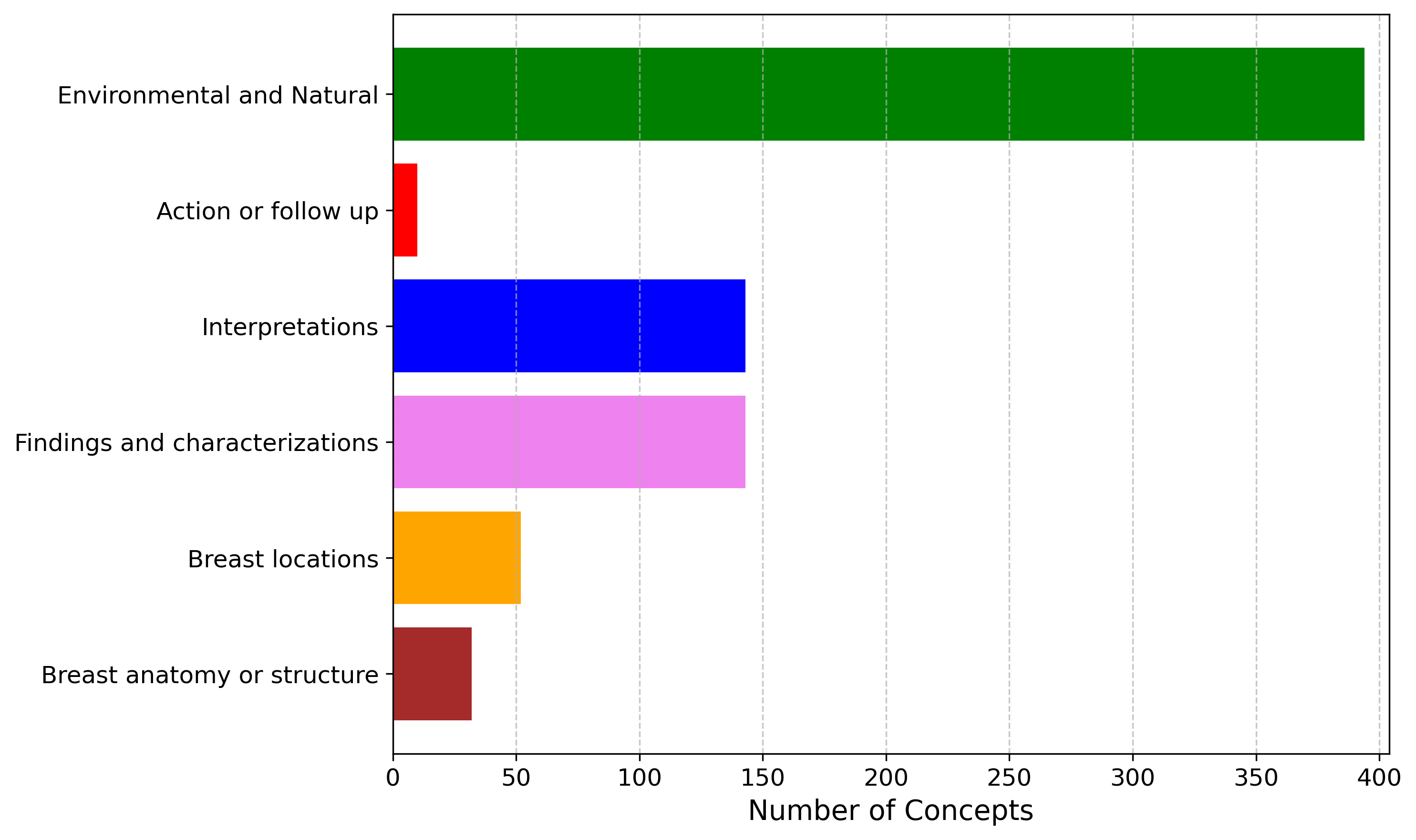}
        \caption{Bar plots of the distribution of the six broad concept categories created in collaboration with radiologists. There are roughly an equal number of mammography and non-mammography concepts.}
        \label{fig:new_broad_cat}
    \end{subfigure}
    \hspace{0.05\textwidth} 
    \begin{subfigure}{0.45\textwidth} 
        \centering
        \includegraphics[width=\linewidth]{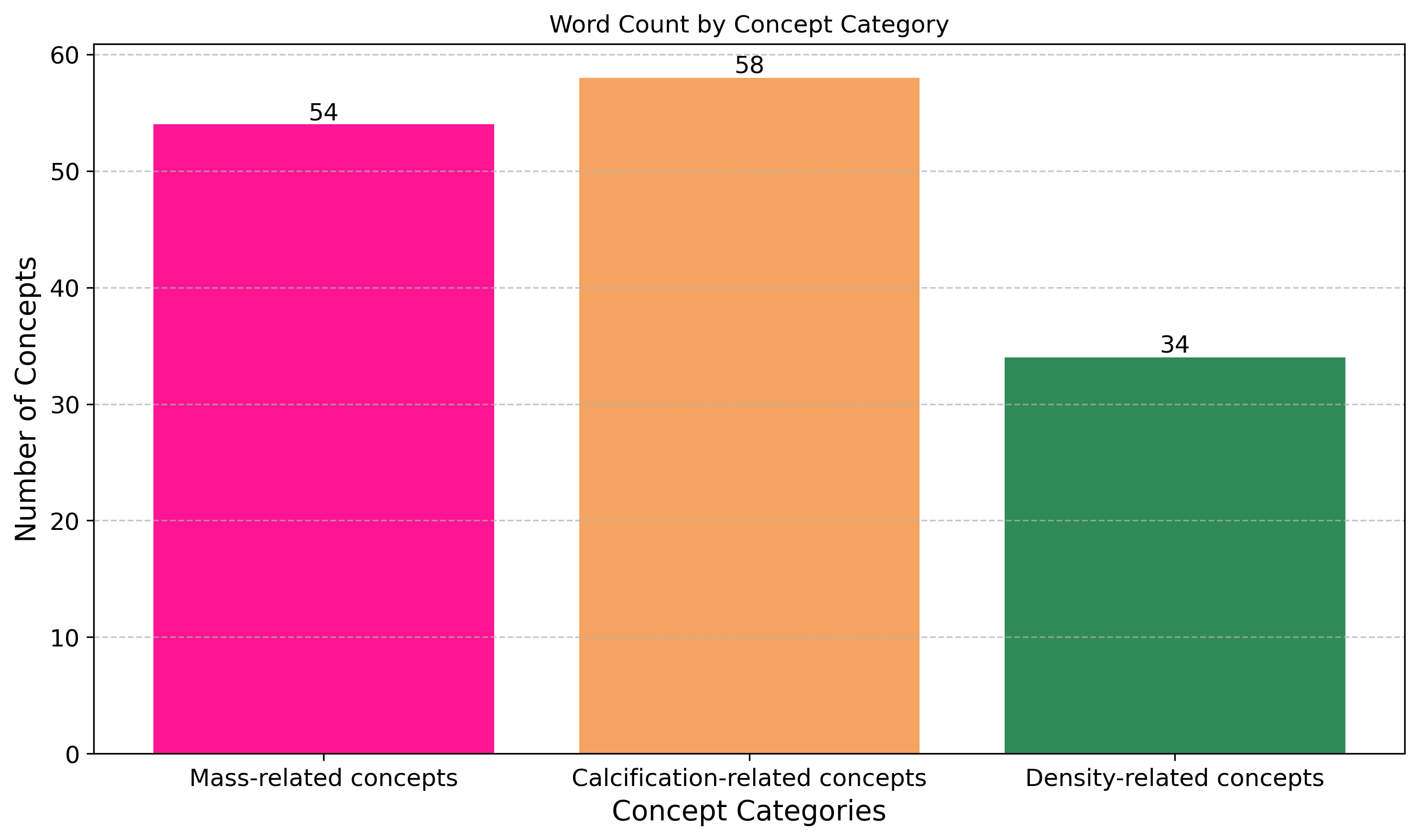}
        \caption{Bar plots highlighting the distribution of concepts related to three specific fine-tuning classification tasks: Mass, Calcification, and Density, present within our concept set.}
        \label{fig:mammo_task_cats}
    \end{subfigure}
    \caption{Subcategorisation of concept set used in the study.}
    \label{fig:figure2}
\end{figure}
In this work, we propose Mammo-CLIP Dissect, a novel framework for layer-wise analysis of mammography-specific concepts learned by Mammo-CLIP. Our approach adapts the CLIP-Dissect framework to the mammography domain by leveraging Mammo-CLIP and introducing a clinically informed, domain-specific concept set. Below, we describe the key components of our methodology, including details on the datasets used, adaptations of Mammo-CLIP and CLIP-Dissect, the curation of the mammography-specific concept set and the fine-tuning process for downstream tasks.

We make our code implementation of Mammo-CLIP dissect along with our concept set and the concept categories we used, publicly available\footnote{\url{https://github.com/Suaiba/Mammo-CLIP-Dissect}}.
\subsection*{Datasets}
Screening mammography typically involves acquiring four images per individual, two images of each breast, left and right, from two standard views. The BI-RADS system is used by radiologists to help categorise and report likelihood of cancer and breast density based upon breast characteristics and presence of clinically relevant findings such as mass and calcifications. 

We utilised the publicly available VinDR-Mammo \cite{Nguyen2023VinDr-Mammo:Mammography} and EMory BrEast imaging Dataset (EMBED) \cite{Jeong2023TheImages} datasets for our experiments. Additionally, the large-scale natural image database, ImageNet \cite{Deng2010ImageNet:Database}, the private University of Pittsburgh Medical Center (UPMC) \cite{Ghosh2024Mammo-CLIP:Mammography}dataset and VinDR-Mammo were employed to pre-train Mammo-CLIP models, which we leveraged.

The UPMC dataset comprises $13,289$ screening mammograms and corresponding text report pairs. This encompasses $25,355$ screening mammograms with subjects having at least one imaging view present. The data was split into $80:20$ ratio for training and test, respectively. The public dataset VinDR-Mammo included $20,000$ mammograms with all four standard views present for each subject. This was divided into training and test sets with $16391$ and $4095$ images, respectively. While no radiology reports are available for VinDR-Mammo, additional attributes such as BI-RADS cancer ($1-5$) and density (A-D) assessments and locations and details on findings (including mass and calcifications) are available for this dataset. These findings were leveraged along with imaging details to generate text reports using templates by Ghosh et. al \cite{Ghosh2024Mammo-CLIP:Mammography}. The pre-processed VinDR-Mammo images, made available by Ghosh et. al \cite{Ghosh2024Mammo-CLIP:Mammography}, of dimensions $ 1520 \times 912$ have been used in our experiments.  The full publicly available EMBED contains $64,564$ mammograms from $23,256$ individuals from different racial backgrounds. Besides mammograms, EMBED contains information on breast density and the number of relevant clinical findings, as well as images with foreign bodies, including implants and clinical markers, which assist us in our analysis of concepts. For some of our experiments, we used a subset of the EMBED dataset with a total of $84$ images comprising $42$ images with implants and $42$ images without implants. We resized these images to be $ 1520 \times 912$.

\subsubsection*{Mammography-Specific Concept Set}
We created a novel mammography-specific concept set in collaboration with two expert radiologists to analyse how different Mammo-CLIP-based models learn clinically relevant concepts. This set also includes non-mammography-related concepts. The mammography-related concepts were derived from multiple sources, including the official BI-RADS taxonomy, findings data, and synthesised radiology reports for the VinDR-Mammo dataset, as well as direct input from radiologists. These concepts consist of both individual words and short, clinically relevant phrases. 

To complement this, non-mammography concepts were collected from the publicly available list\footnote{\url{https://github.com/teresa-sc/concepts_in_ViTs/tree/main/data}} of the $20,000$ concepts most frequently used in the English language, which were categorised into semantic concepts by Dorszewski et al.\cite{Dorszewski2025FromTransformers}. From this list, four semantic concept categories were selected: \emph{Materials}, \emph{Natural}, \emph{Objects}, and \emph{Colours}, to create a non-mammography-related word bank roughly equal in size to the mammography-related concepts. To refine the concept set, we excluded the $100$ most commonly occurring English words, such as articles (\emph{a}, \emph{an}, \emph{is}), and further revised the categories in consultation with the radiologists. This process resulted in a final concept set comprising $763$ unique concepts, which were organised into subcategories and grouped into broader categories for analysis, as shown in Figures \ref{fig:new_broad_cat} and \ref{fig:mammo_task_cats}. 

For the broad concept categories shown in Figure~\ref{fig:new_broad_cat}, we considered two experimental setups: (1) all six broad categories, and (2) a simplified division into mammography-related and non-mammography-related concepts. The six broad categories comprise five mammography-related categories and one non-mammography-related category. The mammography-related concepts are divided into $22$ subcategories reflecting the clinical workflow used by consulting radiologists during mammography analyses. These subcategories capture varying levels of conceptual complexity, progressing from simpler concepts addressed early in the workflow to more complex, interpretative concepts derived later. Ordered by increasing complexity, these are: \emph{Breast anatomy or structures}, \emph{Breast locations}, \emph{Findings and characterizations}, \emph{Interpretations}, and \emph{Action or follow up}. In Figure~\ref{fig:new_broad_cat}, mammography concepts are presented vertically in this order of increasing complexity.  

The sixth broad category, representing non-mammography-related concepts, consists of four subcategories as described above. For experiments comparing mammography-related and non-mammography-related concepts, the 22 mammography subcategories were combined into a single category containing $369$ unique concepts, and the four non-mammography subcategories into another containing $394$ unique concepts.

In addition to these broader categories, we curated task-specific concept categories tailored to three downstream classification tasks: mass classification, calcification classification, and BI-RADS density classification. This resulted in three specialised categories: Mass-related concepts ($73$ concepts), Calcification-related concepts ($79$ concepts), and Density-related concepts ($38$ concepts). The distribution of these task-specific categories is visualised in Figure \ref{fig:mammo_task_cats}.

The proposed concept set and its categorisations are designed to be flexible and can be further customised or refined for additional analyses. This adaptability ensures that the concept set can be tailored to meet the specific needs of future studies or tasks.
\subsection*{Mammo-CLIP-dissect framework}
An overview of the implemented Mammo-CLIP Dissect framework is presented in Figure \ref{fig:overview}. The framework closely follows the CLIP-dissect implementation and incorporates four primary components: (1) a set of probing images, $D_{probe}$; (2) a target model, $F_{target}$, whose neurons are to be labelled---this can be any DL vision model; (3) a VLM as the dissector model, $F_{dissector}$, such as Mammo-CLIP, where the text and image encoders of the $F_{dissector}$ model embed the probing images and facilitate the dissection of the $F_{target}$ model; and (4) a predefined set of concepts, $\mathcal{C}$, with $M$ concepts. This concept set is used to assign labels to the neurons of $F_{target}$. 

These components are leveraged for neuron labelling in Mammo-CLIP-dissect using three stages:

\begin{itemize}

    \item Each probing image $\mathbf{x_{i}} \in D_{probe} \subset \mathbb{R}^{Channel \times Height \times Width}$ from the probing set $D_{probe}$ is embedded using the image encoder of the dissector model, while the concepts $\mathbf{t_j}$ from the concept set $\mathcal{C}$ are embedded using its text encoder. The inner product between image embedding $\mathbf{I_i}$ of probe image $\mathbf{x_i}$, and concept embedding $\mathbf{T_j}$ of concept $\mathbf{t_j}$, is computed to form a concept-activation matrix, $\mathbf{P} \in \mathbb{R}^{N \times M}$, where N and M are the number of image and text embeddings, respectively. $\mathbf{P}$ quantifies the similarity between embeddings of the images and concepts, where $\mathbf{P}_{i,j} = \langle \mathbf{I_i} \cdot \mathbf{T_j} \rangle$. This matrix acts as a look-up dictionary when comparing neuron activations to concept activations in the next steps.
    \item For each neuron, $k$, in $F_{target}$, its mean activation map, $A_k(\mathbf{x_i}) \in \mathbb{R}$, is recorded for each probing image $\mathbf{x_i} \in D_{probe}$. 
    These neuron activations across all probing images are collected in the activation vector $\mathbf{q_k} = \begin{bmatrix} A_k(\mathbf{x_1}), \dots, A_k(\mathbf{x_N}) \end{bmatrix}^T \in \mathbb{R}^N$. $\mathbf{q_k}$ is then used to match the neuron activations to the concept activations over the same probing dataset.
    \item The label assigned to each neuron, $k$, by defining by the most similar concept $t_m$ to $k$ as per the activation vector $\mathbf{q_k}$ comprising of the activations $A_k$. We do this by comparing \(\mathbf{q_k}\) with the concept–activation matrix using a similarity function \(\text{sim}(t_m, \mathbf{q_k})\). We then define
\[
\tilde{m} = \arg\max_{m} \text{sim}(t_m, \mathbf{q_k}),
\]
where \(\tilde{m}\) is the index of the most similar concept in the concept set \(\mathcal{C}\) (with \(M\) concepts). The concept label assigned to neuron \(k\) is thus \(t_{\tilde{m}}\), the concept corresponding to index \(\tilde{m}\).
      Soft Weighted Pointwise Mutual Information (SoftWPMI) \cite{Oikarinen2022CLIP-Dissect:Networks,Wang2020TowardsImages} was the similarity function selected in this work based on the original CLIP-Dissect implementation by Oikarinen et. al. \cite{Oikarinen2022CLIP-Dissect:Networks}. 
    SoftWPMI is computed by computing the mutual information (MI) between the set of images 
    $D^{\mathbf{q_k}}_{probe} = \{\mathbf{x_i} \in D_{probe}: A_k(\mathbf{x_i})\}$ is among the top-Z largest entries of $\mathbf{q_k}$ 
    and the concept $t_m$. SoftWPMI is formulated as:
    \begin{equation}
      \text{sim}(t_m, \mathbf{q_k}) \triangleq \text{SoftWPMI} (t_m, \mathbf{q_k}) =  \log \mathbb{E} [p (t_m | D^{\mathbf{q_k}}_{probe})] - \lambda \log p(t_m)
    \end{equation}
\end{itemize}
Here, $\lambda$ is a regularising parameter. The term $\log \mathbb{E} [p (t_m | D^{\mathbf{q_k}}_{probe})] = \log (\prod_{\mathbf{x_i} \in D_{probe}} [1 + p (\mathbf{x_i} \in D^{\mathbf{q_k}}_{probe}) (p(t_m | \mathbf{x_i}) - 1)])$, further details and derivation of this are available in the work by Oikarinen et al \cite{Oikarinen2022CLIP-Dissect:Networks}. The SoftWPMI similarities between each neuron and concept are collected in the matrix $\mathbf{S}$.

This process enables the automatic assignment of the most relevant concept labels to neurons, providing insights into the concepts learned by the target model, $F_{target}$.

\subsubsection*{Mammo-CLIP models}
Mammo-CLIP architecture consists of an image encoder, $E_I$ and a text encoder, $E_T$, which are pre-trained to capture domain-specific features, using a contrastive loss to align image and text representations. 
 For $N$ pairs of images $\mathbf{x_{i}}$ and text $\mathbf{t_i}$, Mammmo-CLIP computes image embeddings, $\mathbf{I_i}=E_I(\mathbf{x_{i}})$, and text embeddings, $\mathbf{T_i}=E_T(\mathbf{t_i})$. The training objective is to maximise the SoftWPMI similarity between matching pairs $(\mathbf{I_i}, \mathbf{T_i})$ and minimise the SoftWPMI similarity between non-matching pairs $(\mathbf{I_i}, \mathbf{T_j})$, where $j \neq i$.
 
In our approach, we adopted Mammo-CLIP with the CNN EfficientNet-B5 \cite{Tan2019EfficientNet:Networks} as the image encoder and the transformer-based BioClinicalBERT \cite{Alsentzer2019PubliclyEmbeddings} as the text encoder as per the original implementation by Ghosh et al. \cite{Ghosh2024Mammo-CLIP:Mammography}. 

\textbf{$\mathbf{F_{general}}$:} This general, non-mammography specific setup involves an EfficientNet-B5 backbone pre-trained on the ImageNet dataset, and the BioClinicalBERT encoder pre-trained on general biomedical text to encode radiology reports.

\textbf{$\mathbf{F_{mammo}}$:} To specialise Mammo-CLIP for screening mammography data, this general setup was further pre-trained on paired screening mammograms and radiology text reports by Ghosh et al. \cite{Ghosh2024Mammo-CLIP:Mammography}. Specifically, a combination of the private UPMC dataset (with its associated radiology reports) and the public VinDR dataset was used to fine-tune the encoders in Mammo-CLIP. For our experiments, we employed the publicly available weights from this mammography-specific pre-trained setup \footnote{\url{https://github.com/batmanlab/Mammo-CLIP}}, which we refer to as $F_{mammo}$.

The image encoder of $F_{mammo}$ was further tailored for various downstream mammography classification tasks by adding a linear classification layer and fine-tuning the overall architecture. These tasks included: \begin{itemize}
    \item BI-RADS density classification,
    \item BI-RADS cancer classification,
    \item binary mass classification and 
    \item binary suspicious calcification classification.
\end{itemize}
We refer to this collection of fine-tuned Mammo-CLIP image encoder-based classifier setups as $F_{classifier}$.

We explored three main Mammo-CLIP dissect configurations presented in Table \ref{tab:mammo_clip_dissect}.
\begin{table}[h!]
\centering
\resizebox{\columnwidth}{!}{
\begin{tabular}{|c|c|c|>{\centering\arraybackslash}p{8cm}|c|}
\hline
\textbf{Setup} & \textbf{\( F_{target} \)} & \textbf{\( F_{dissector} \)} & \textbf{Purpose} & \textbf{D\_probe} \\ \hline
G-Mammo-CLIP Dissect & \( F_{general} \) & \( F_{general} \) & Evaluates the performance of a general Mammo-CLIP without domain-specific pre-training. & VinDR-Mammo \\ \hline
M-Mammo-CLIP Dissect & \( F_{mammo} \) & \( F_{mammo} \) & Assesses the benefits of mammography-specific training on Mammo-CLIP. & VinDR-Mammo \\ \hline
C-Mammo-CLIP Dissect & \( F_{classifier} \) & \( F_{mammo} \) & Evaluates task-specific classifier performance while leveraging domain knowledge in \( F_{mammo} \). & VinDR-Mammo \\ \hline
\end{tabular}}
\caption{Summary of Mammo-CLIP Dissect Configurations.}
\label{tab:mammo_clip_dissect}
\end{table}
In all three experimental setups, the dissector network $F_{dissector}$ is Mammo-CLIP, incorporating both its image and text encoders and the $D_{probe}$ is VinDR-Mammo. 
By contrast, the target network $F_{target}$ employs only the image encoder of Mammo-CLIP. 
The primary distinction between the G-Mammo-CLIP Dissect and M-Mammo-CLIP Dissect configurations lies in the data used for model training. Each setup is designed to address a distinct research question:
\begin{itemize}
    \item \textbf{G-Mammo-CLIP Dissect},evaluates how general, non-mammography specific Mammo-CLIP trained on ImageNet and general medical text performs when applied to mammography, thereby establishing a baseline without domain-specific pretraining. Within this setup, the $F_{target}$ is the image encoder of $F_{general}$. The $F_{dissector}$ in G-Mammo-CLIP Dissect includes both the image and text encoders of $F_{general}$.
    \item \textbf{M-Mammo-CLIP Dissect} assesses the effect of domain-specific training by using Mammo-CLIP models pretrained on mammography images (UPMC and VinDR-Mammo) and corresponding radiology reports. In this setup, $F_{target}$  corresponds to the image encoder of $F_{mammo}$. The $F_{dissector}$ in M-Mammo-CLIP Dissect consists of both encoders of $F_{mammo}$.
    \item \textbf{C-Mammo-CLIP Dissect} examines task-specific fine-tuning by using $F_{classifier}$ as the $F_{target}$. $F_{classifier}$ is an image classifier built upon the image encoder of $F_{mammo}$. This third setup helps us assess the impact of fine-tuning for downstream mammography tasks on the concepts learned.
\end{itemize}
\subsection*{Concept-Based Analysis}
We base our concept analyses on Mammo-CLIP models with an EfficientNet-B5 image encoder as discussed. Specifically, we focus on three representative convolutional layers 
corresponding to the early, middle, and late stages of the model, respectively (Figure \ref{fig:effb5_standard}). At each of these layers, we evaluate the concepts captured by computing similarity scores between neuron activations and concept representations. 

To determine which concepts are most relevant to neurons within a given layer, we adopt an adaptive, model-specific thresholding scheme inspired by Dorszewski et al.~\cite{Dorszewski2025FromTransformers}. 
For each layer \(l \in \{1,2,\dots,L\}\) of the target model \(F_{\text{target}}\), we compute a threshold \(\tau_l\) equal to the mean similarity score between neuron activations and concept representations at that layer: $\tau_l = \text{mean}(S_l)$, where \(S_l\) denotes the SoftWPMI similarity scores between all neurons and all concepts in layer \(l\).

When comparing two models, we define the threshold at each layer as the maximum of the two models’ mean similarity scores to ensure fair comparison: $\tau_l = \max\bigl(\tau^{\text{Model}_1}_l,\,\tau^{\text{Model}_2}_l\bigr),$
where $\tau^{\text{Model}_i}_l$ is the mean similarity score at layer \(l\) for Model \(i\). 
For example, when comparing G-Mammo-CLIP Dissect and M-Mammo-CLIP Dissect using the same \(D_{\text{probe}}\), \(\tau_l\) is determined by the higher mean similarity at that layer across the two models.

A concept is considered encoded at a layer if its similarity score with the neuron activations is greater than or equal to \(\tau_l\). 
This adaptive thresholding accounts for model-specific pretraining differences and allows fairer comparisons across layers. 
Consequently, the number of neurons considered activated varies by layer depending on the threshold.

By examining the number and types of concepts activated at each layer, we can uncover and compare the representational patterns learned by different mammography models at different stages.  Additionally, comparing these thresholds across models offers insight into differences in representations and selectivity throughout the network.

\subsection*{Implementation details}
 The training process for $F_{mammo}$ models used in our framework involved 10 epochs using an AdamW \cite{Loshchilov2017DecoupledRegularization} optimiser and a contrastive loss to align image and text representations. We set $\lambda = 1$ for the SoftWPMI similarity function $\text{sim}(t_m, \mathbf{q_k})$, following Oikarinen et al.~\cite{Oikarinen2022CLIP-Dissect:Networks}. The parameter $Z$ was chosen based on the size of the probe dataset $D_{\text{probe}}$: for VinDR-Mammo, we used $Z = 100$ as in Oikarinen et al.~\cite{Oikarinen2022CLIP-Dissect:Networks}, whereas for the smaller EMBED subset of $D_{\text{probe}}$ containing $84$ images, we set $Z = 84$.

Building on these pre-trained $F_{mammo}$ models, $F_{classifier}$ models were implemented by fine-tuning the $F_{mammo}$ image encoders for four downstream mammography classification tasks. The fine-tuning setup closely followed the original configuration described by Ghosh et. al. \cite{Ghosh2024Mammo-CLIP:Mammography}. Specifically, a single fully connected linear layer was added on top of $F_{mammo}$, and the overall architecture was then trained for the specified classification task. The fine-tuning process was conducted for 30 epochs with a learning rate of $(5 \times 10^{-5})$, a batch size of $8$, and the AdamW \cite{Loshchilov2017DecoupledRegularization} optimiser. A LinearWarmupCosineAnnealingLR scheduler was employed to adjust the learning rate dynamically, with warmup steps set to $10$. For the classification tasks, different loss functions were used depending on the task type. For binary classification tasks (mass and calcification), a weighted binary cross-entropy loss was optimised. The positive class weights were determined based on the class imbalance in the dataset. For multi-class classification tasks (BI-RADS density and cancer), a standard cross-entropy loss was used. The evaluation metrics also varied based on the task type. For binary classification tasks, binary AUC was used as the primary evaluation metric. For multi-class classification tasks, multi-class accuracy and F1 scores were used.

All code was implemented on PyTorch \cite{Paszke2019PyTorch:Library}. For all neuron-labelling experiments, we utilised NVIDIA RTX 3090 and A6000 GPUs. The neuron labelling experiments took roughly $1.5$ hours. For the fine-tuning experiments, we used  AMD Instinct MI210 GPUs. The training time for fine-tuning the classifier models ranged from $44$ to $70$ hours.

\begin{figure}[h]
    \centering
    \begin{subfigure}[t]{0.5\linewidth} 
        \centering
        \includegraphics[width=0.9\linewidth]{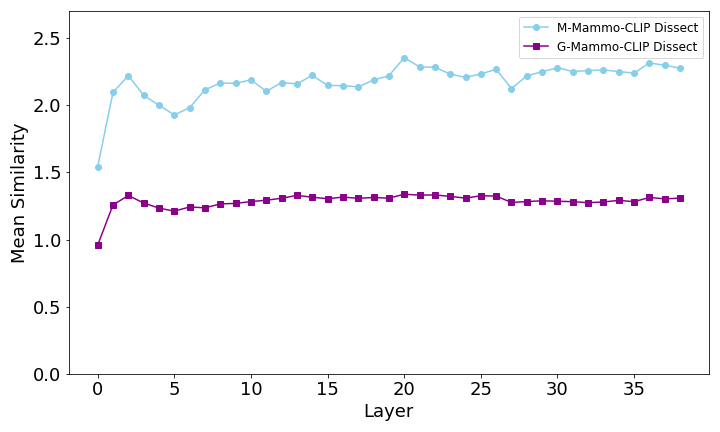}
        \caption{Evolution of mean layer similarities.}
        \label{fig:global_new_means_ft_extract}
    \end{subfigure}
    \vskip 1em 
    \begin{subfigure}[t]{0.5\linewidth} 
        \centering
        \includegraphics[width=\linewidth]{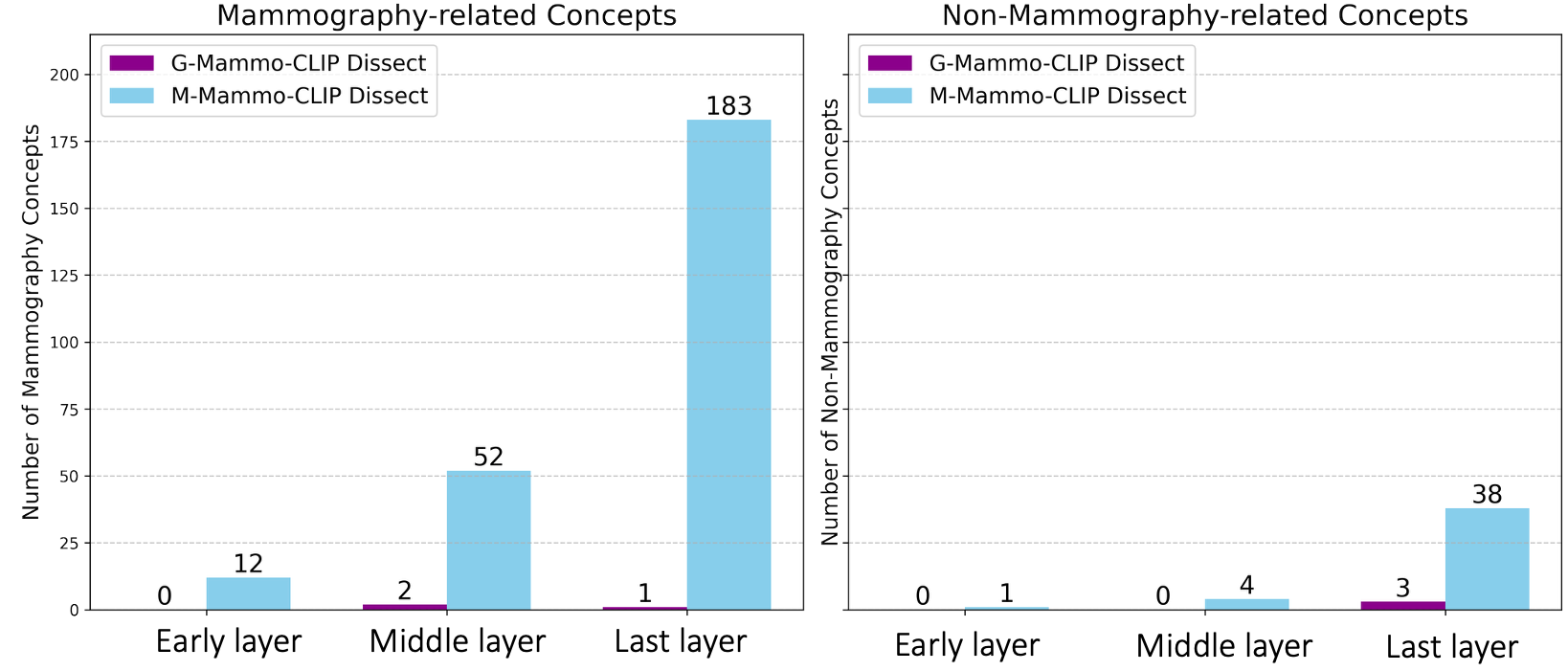}
        \caption{Comparison of occurrences of unique mammography and non-mammography-related concepts.}
        \label{fig:global_side_by_side_binary_comp}
    \end{subfigure}
    \hfill
    \begin{subfigure}[t]{0.48\linewidth} 
        \centering
        \includegraphics[width=\linewidth]{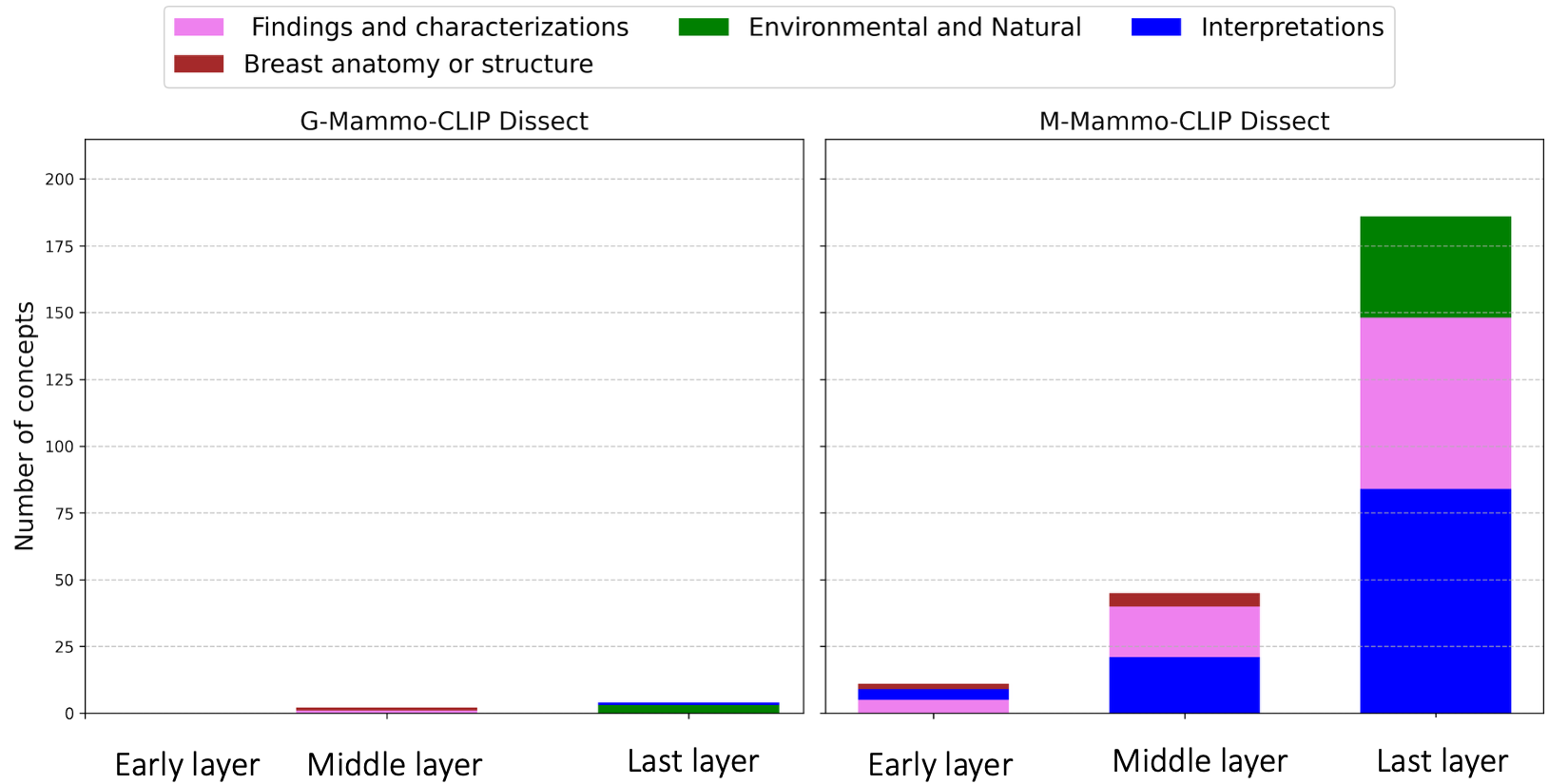}
        \caption{The top three broad concept categories learned by the two models.}
        \label{fig:global_top3_broad_ft_comp}
    \end{subfigure}
    \caption{Comparison of G-Mammo-CLIP Dissect (pretrained on ImageNet) and M-Mammo-CLIP Dissect (pretrained on mammography) with VinDR-Mammo as the probe. Mean layer-specific similarity thresholds have been applied in these plots. (a) Evolution of mean layer similarities, used to determine threshold, $\tau$, across G-Mammo-CLIP Dissect and M-Mammo-CLIP Dissect. (b) Occurrences of mammography and non-mammography-related concepts across layers of G-Mammo-CLIP Dissect and M-Mammo-CLIP Dissect. (c) Stacked bar plots visualising the top three broad concept categories learned by G-Mammo-CLIP Dissect on the left and M-Mammo-CLIP Dissect on the right.}
    \label{fig:global_combined_figures}
\end{figure}
\section*{Results}
In this section, we present our results for the three research questions posed. 
\subsection*{\textbf{RQ1: Do CNNs trained on mammography data learn more mammography-specific concepts compared to models trained on non-mammography data?}}

Within this section, we compare how G-Mammo-CLIP Dissect, which has been trained on ImageNet and M-Mammo-CLIP Dissect, which has been trained on mammography data, learn in terms of concepts. 

\begin{figure}[!ht]
    \centering
\includegraphics[width=0.8\linewidth]{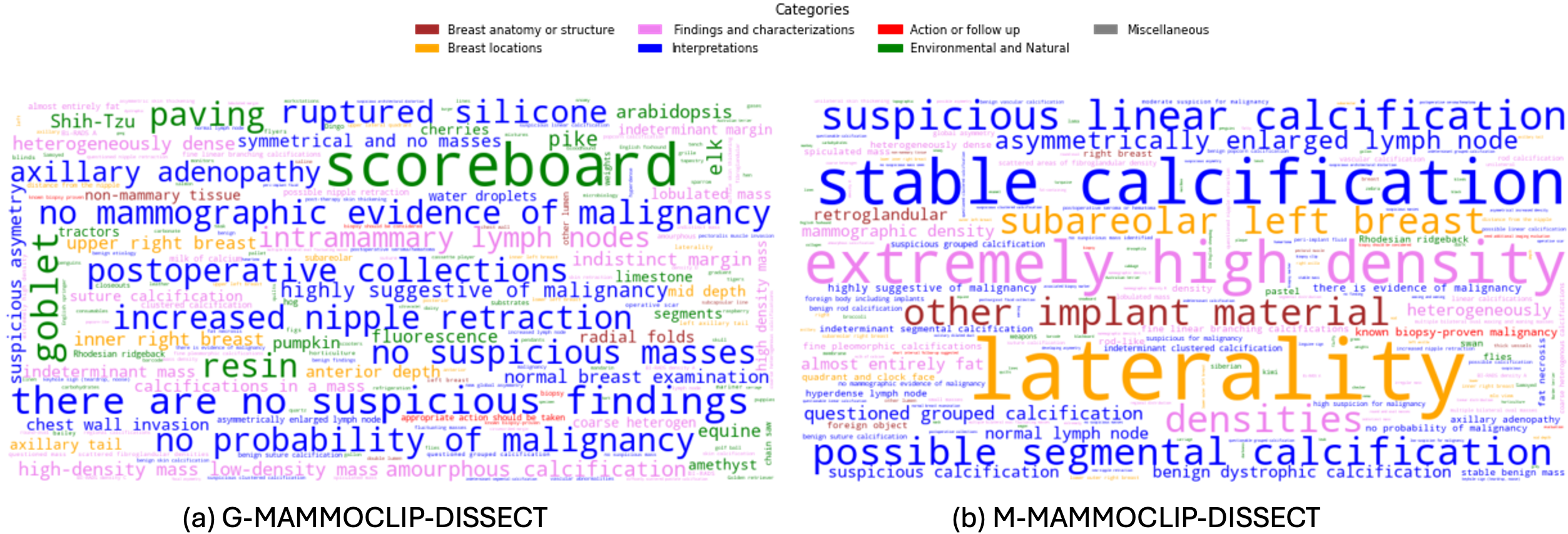}
    
    \caption{Comparison of wordclouds of unique concepts for G-Mammo-CLIP Dissect and M-Mammo-CLIP Dissect at the last layer. The size of words is proportional to their similarity value with neurons and the colour of the broad category to which they belong.}
    \label{fig:comparison_last_wordcloud}
\end{figure}

First, we compare the mean similarity scores between neurons and concepts, used as thresholds, $\tau$, at each layer within both $F_{target}$ models. This analysis is presented in Figure \ref{fig:global_new_means_ft_extract}. From this figure, we can observe that M-Mammo-CLIP Dissect consistently has higher $\tau$ values than G-Mammo-CLIP Dissect.  Next, we analyse the number of unique concepts captured, surpassing the layer-specific threshold $\tau$ across three representative layers. Figures \ref{fig:global_side_by_side_binary_comp} and \ref{fig:global_top3_broad_ft_comp} summarise these findings for G-Mammo-CLIP Dissect and M-Mammo-CLIP Dissect.  

\begin{figure}[!ht]
    \centering
    \includegraphics[width=0.7\linewidth]{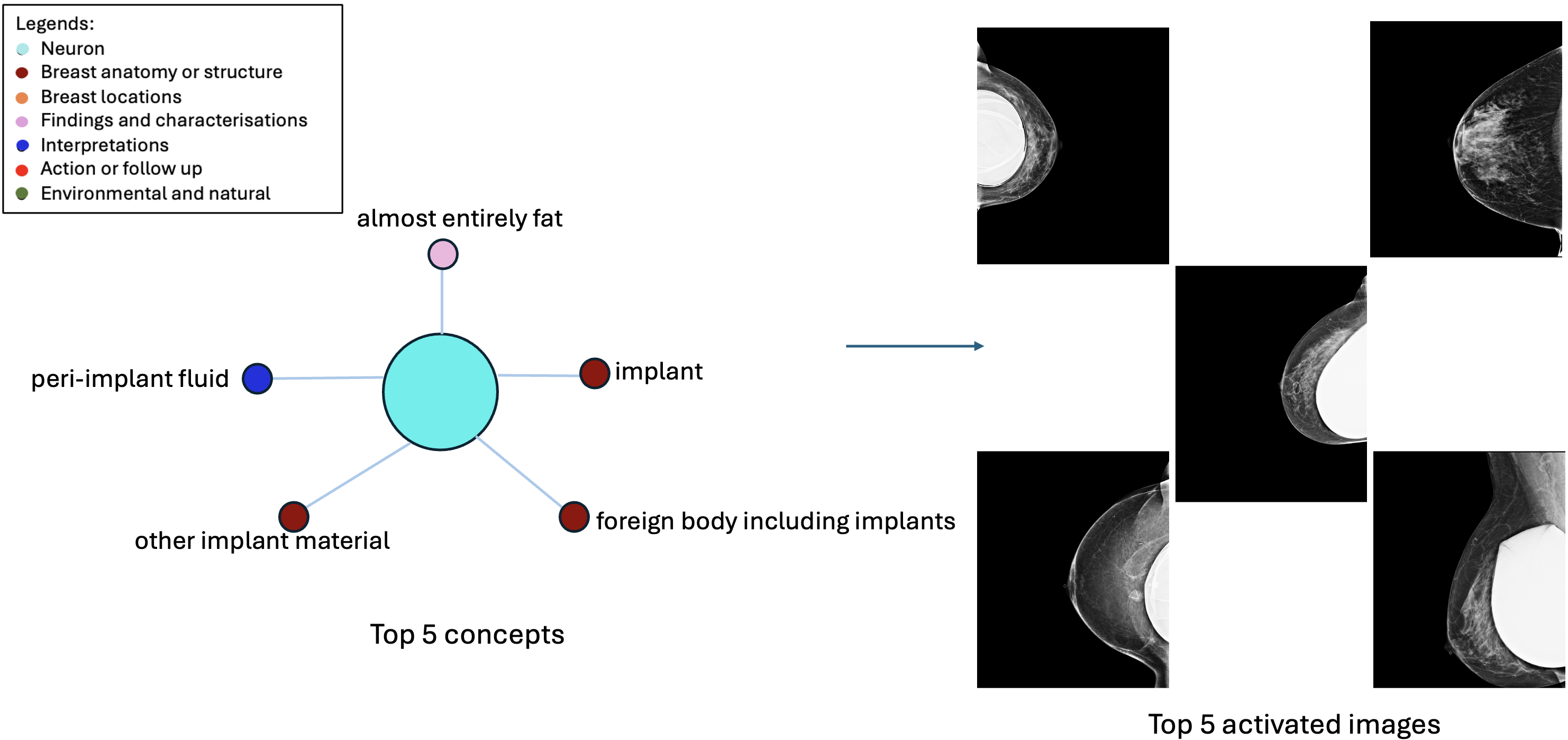}
    \caption{Top five activated concepts and images of neuron $451$, at the last layer of M-Mammo-CLIP Dissect, which meets the global mean threshold. In this case, the probe is a subset of EMBED.}
    \label{fig:implant451}
\end{figure}
\begin{figure}[!ht]
    \centering
    \includegraphics[width=\linewidth]{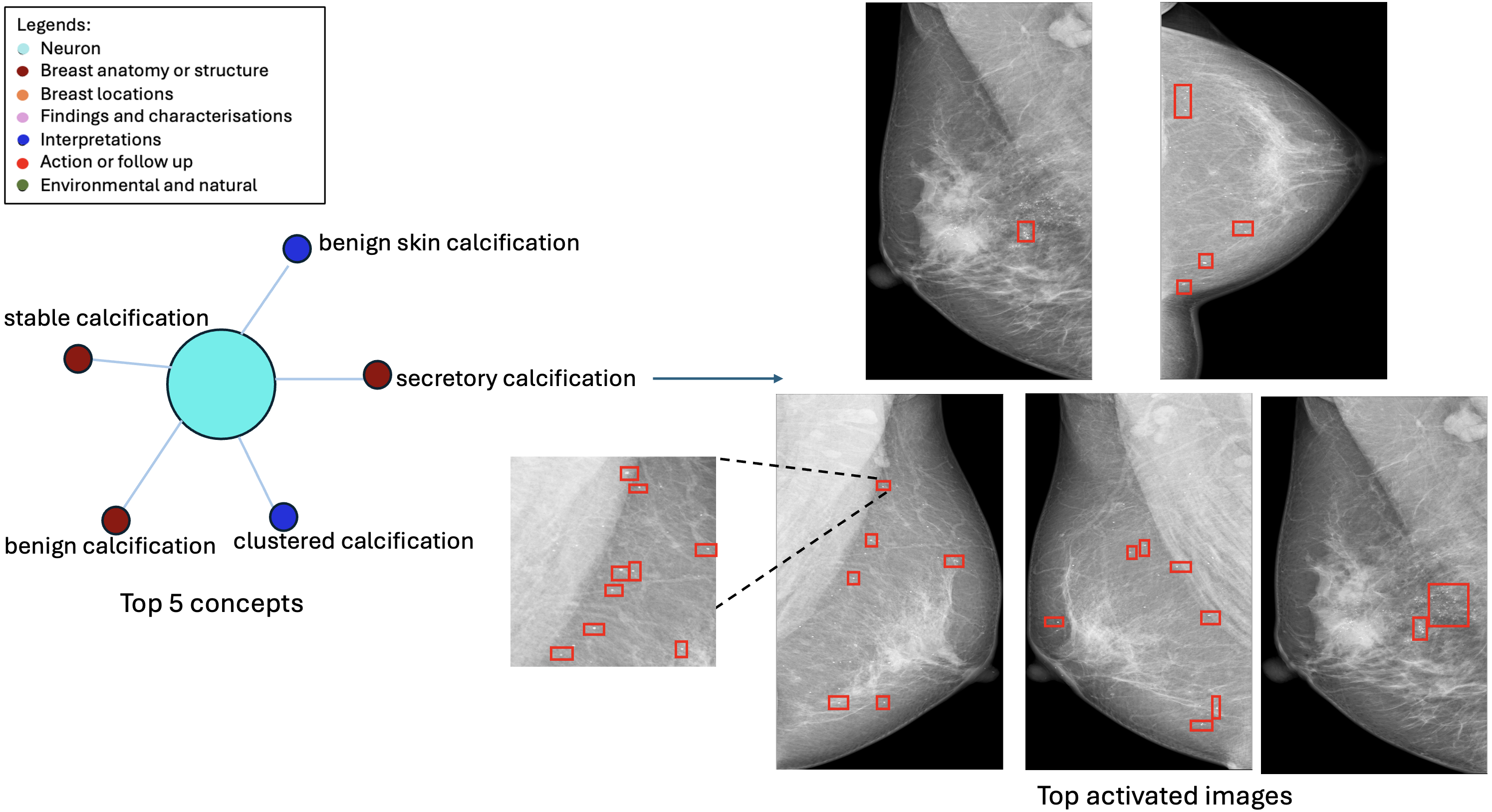}
    \caption{Top five activated concepts and images of neuron $242$, at the last layer of M-Mammo-CLIP Dissect, which meets the global mean threshold. In this case, the probe is VinDR-Mammo.}
    \label{fig:calc242}
\end{figure}

Expanding on these results, Figure \ref{fig:global_side_by_side_binary_comp} presents the number of unique mammography-related and non-mammography-related concepts captured by G-Mammo-CLIP Dissect and M-Mammo-CLIP Dissect. M-Mammo-CLIP Dissect learns substantially more concepts overall, including both mammography-specific and general ones. Importantly, mammography-related concepts dominate within M-Mammo-CLIP Dissect.

We also examined which broad categories of concepts ranked among the top three most strongly associated with each model in Figure \ref{fig:global_top3_broad_ft_comp}. For M-Mammo-CLIP Dissect, the categories \emph{Findings and Characterizations} and \emph{Interpretations} consistently emerged, both highly relevant for mammography analysis. In contrast, G-Mammo-CLIP Dissect showed no stable pattern: its middle layers focused on mammography-related categories, but the last layer shifted toward a mixture of mammography and non-mammography categories. 

To visualise the concepts themselves, Figure \ref{fig:comparison_last_wordcloud} presents word clouds of the last-layer concepts for both $F_{target}$ models. Concepts are coloured by category, with word size reflecting activation strength. M-Mammo-CLIP Dissect clearly emphasises mammography-related concepts, while G-Mammo-CLIP Dissect places greater focus on concepts from the \emph{Environmental and Natural} category.

Taken together, these findings are consistent with expectations that models trained on mammography data will exhibit stronger, more stable alignment with mammography-relevant concepts and categories than models trained on non-mammography datasets.


We explore other combinations of $D_{probe}$ and $F_{dissector}$ in Appendix I to further highlight the flexibility of the Mammo-CLIP Dissect framework.

The analyses in Figures \ref{fig:global_combined_figures} and \ref{fig:comparison_last_wordcloud} provide a holistic view of the concepts learned in $F_{target}$ models within G-Mammo-CLIP Dissect and M-Mammo-CLIP Dissect. However, individual neurons can be analysed to identify specific behaviour. Prior work by Bykov et al. \cite{Bykov2022DORA:Networks} demonstrated that neurons can react to distinct artefacts. Extending this approach to mammography, we can identify neurons that detect clinically meaningful patterns or reflect potential biases.

 One such example involves breast implants; we chose breast implants as they are visually distinct and easily interpretable, even for non-experts in mammography analysis. Figure \ref{fig:implant451} shows neuron $451$ from the last layer, evaluated on a subset of EMBED as $D_{probe}$ containing $42$ images with implants and $42$ without. The top five activated concepts and images for this neuron reveal a strong implant-related signal: four of the top five concepts explicitly reference implants, and four of the top five images indeed contain implants. This suggests that individual neurons can specialise in clinically meaningful mammography concepts.

A second example is presented in Figure \ref{fig:calc242} using the VinDR-Mammo test set probe. Neuron $242$ exhibits the top five concepts, all of which contain the term \emph{"calcification"}. Correspondingly, each of the top five activated images displays multiple calcifications, which appear as bright spots on the mammograms; selected regions are highlighted with red boxes. Since calcifications are a critical marker in breast cancer screening, this result underscores the model’s ability to capture diagnostically important visual features at the neuron level.

\subsection*{\textbf{RQ2: How does fine-tuning for mammography-relevant tasks, such as density classification, affect the learning of mammography-specific concepts?}}

In this section, we examine how fine-tuning influences the mammography-specific concepts captured by Mammo-CLIP. 
We compare M-Mammo-CLIP Dissect (trained on mammography data) with C-Mammo-CLIP Dissect (fine-tuned on a specific mammography classification task) to identify which concepts are gained or lost during fine-tuning. 
This focus is motivated by Dorszewski et al. \cite{Dorszewski2025FromTransformers}, who demonstrated that fine-tuning can substantially change the internal concepts learned by models, potentially altering their downstream performance. 
By investigating these effects, we aim to better understand how task-specific adaptation impacts concept-level representations and their alignment with clinically relevant features.


\begin{table}[h!]
\centering
\caption{The four fine-tuning classification tasks and their corresponding C-Mammo-CLIP Dissect variants. For each variant, the corresponding task-specific $F_{target}$ name is reported as well as the classification performance. Here, $F_{dissector}$ is $F_{mammo}$ in all cases.}
\label{tab:cmammo_variants}
\begin{tabular}{|c|c|c|c|}
\hline
\textbf{Variant} & \textbf{Fine-tuning Task} & \textbf{$F_{target}$} & \textbf{Classification performance}\\
\hline
C1 & BI-RADS density classification & $F_{classifier}^{C1}$ & $0.860$ (Accuracy), $0.576$ (F1)\\
C2 & BI-RADS cancer classification  & $F_{classifier}^{C2}$ & 0.795 (Accuracy), 0.523 (F1) \\
C3 & Binary mass classification     & $F_{classifier}^{C3}$ & 0.874 (AUC) \\
C4 & Binary suspicious calcification classification & $F_{classifier}^{C4}$ & 0.975 (AUC)\\
\hline
\end{tabular}
\label{Table:res}
\end{table}

Specifically, we analyse four variants of C-Mammo-CLIP Dissect, each corresponding to a distinct mammography classification task as reported in Table \ref{Table:res}. The four downstream classification tasks investigated are—BI-RADS density, binary mass, binary calcification, and BI-RADS cancer—along with their corresponding C-Mammo-CLIP Dissect variants. Each variant uses a task-specific \(F_{target}\) (the classifier) while sharing the same \(F_{dissector}\) (\(F_{mammo}\)). In all cases, \(F_{target}\) consists of the \(F_{mammo}\) image encoder with an added linear classification layer, fine-tuned for the respective task. For example, the BI-RADS density classification variant (C1) employs \(F_{classifier^{C1}}\) as its \(F_{target}\), fine-tuned specifically for that task. Performance metrics are reported per task: multi-class accuracy and F1 for BI-RADS density and BI-RADS cancer, and binary AUC for the mass and calcification tasks. Across all tasks, the fine-tuned models demonstrate strong performance on both accuracy and AUC metrics.
\begin{figure}
 \centering
        \includegraphics[width=0.45\linewidth]{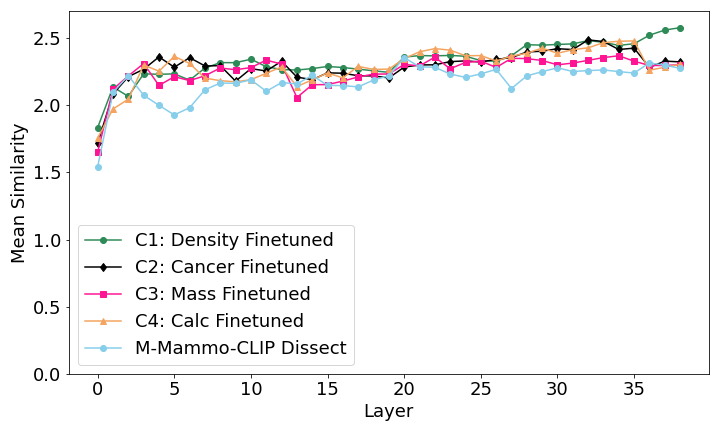}
        \caption{Evolution of mean layer similarities, used as thresholds, across three different $F_{classifier}$ models. Each model was fine-tuned for a different downstream task.}
        \label{fig:new_means_downstream}
\end{figure}

Figure \ref{fig:new_means_downstream} shows the mean similarity thresholds ($\tau$) across $F_{target}$ for M-Mammo-CLIP Dissect and the four variants of C-Mammo-CLIP Dissect. Across all layers, the C-Mammo-CLIP Dissect variants exhibit higher $\tau$ values than M-Mammo-CLIP Dissect, although the overall trend across layers remains consistent. 
\begin{figure}[!ht]
    \centering
    \begin{subfigure}{0.3\textwidth}
        \centering
        \includegraphics[width=\linewidth]{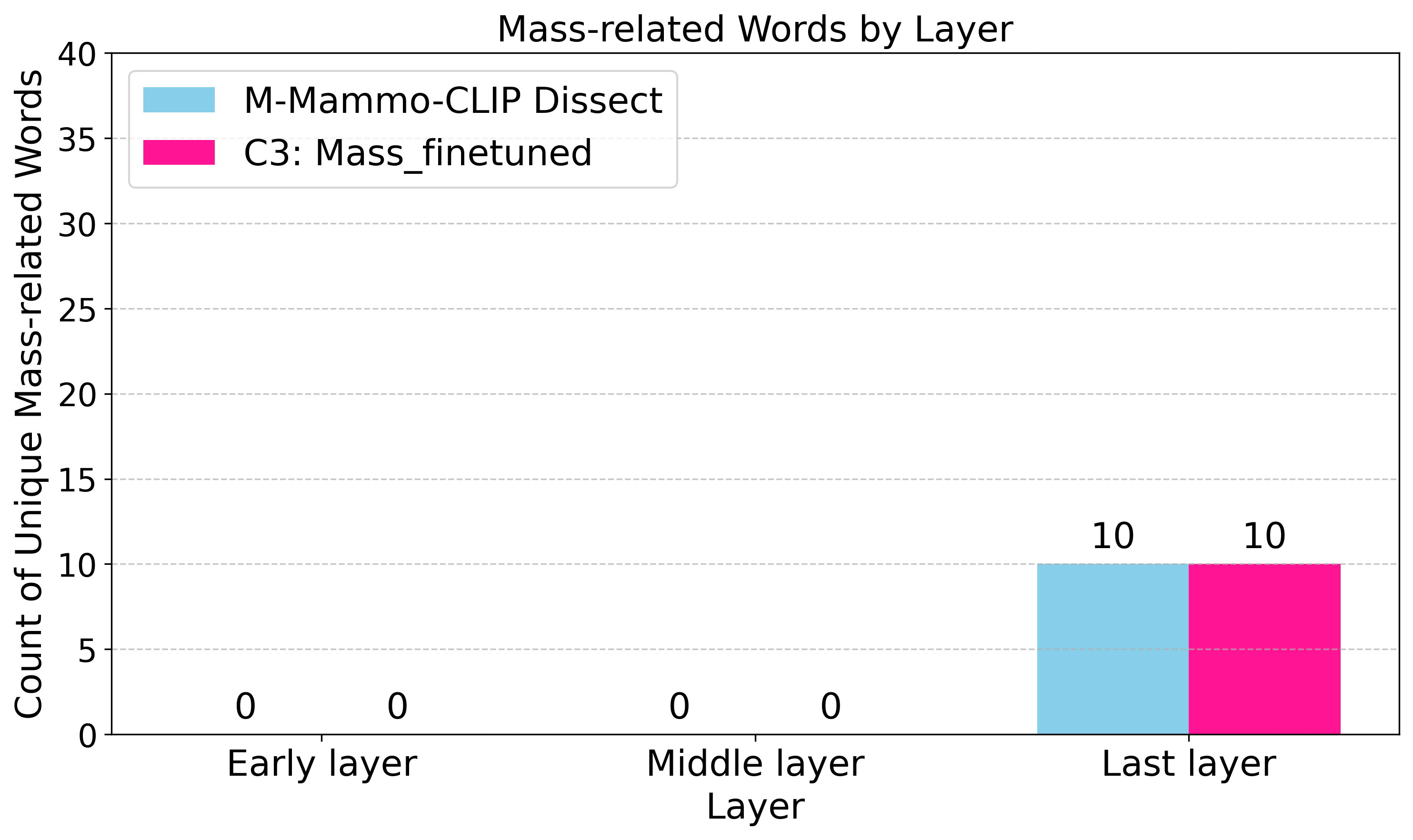}
        \caption{Mass related concepts}
        \label{fig:mass_concepts}
    \end{subfigure}
    \hspace{0.03\textwidth}
    \begin{subfigure}{0.3\textwidth}
        \centering
        \includegraphics[width=\linewidth]{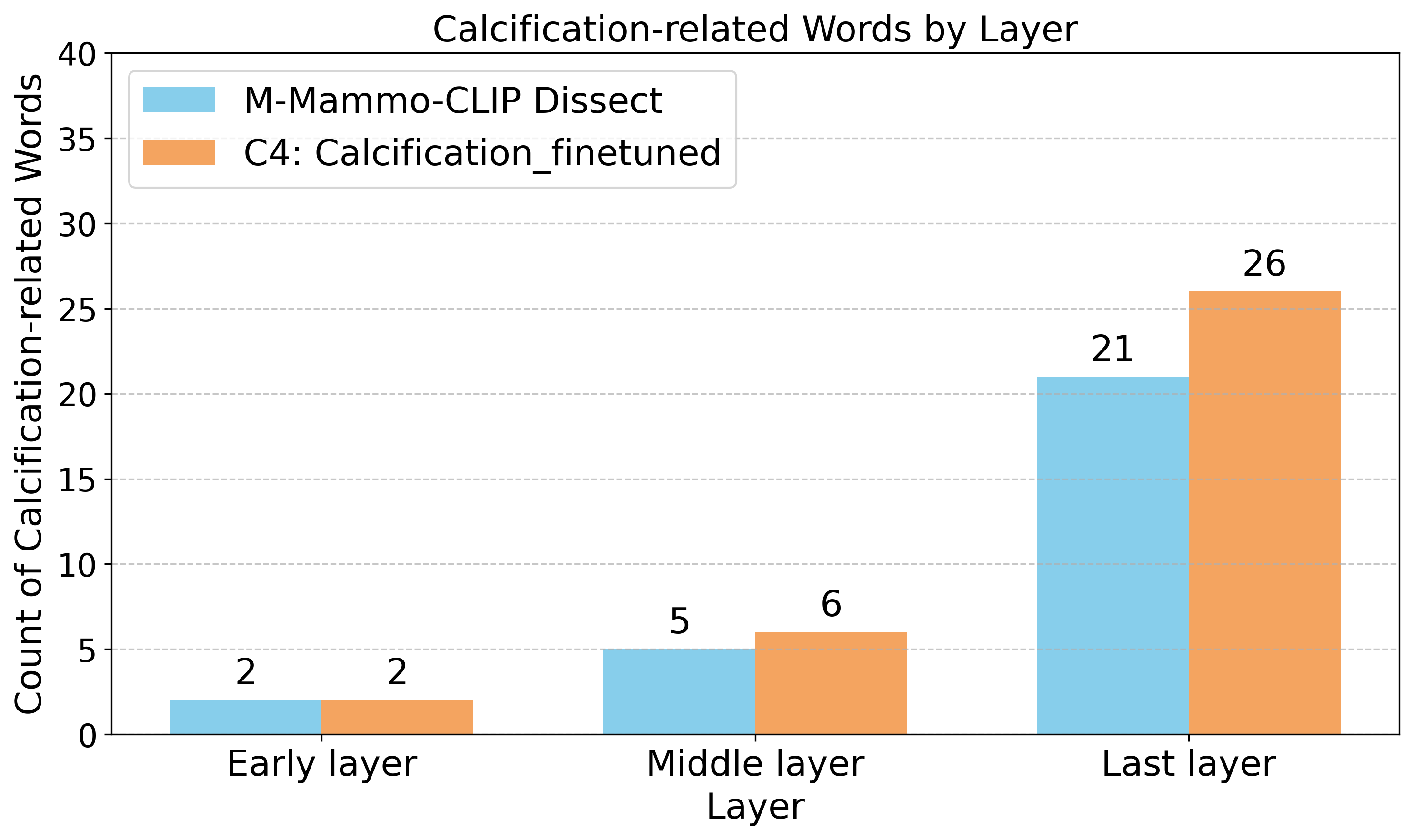}
        \caption{Calcification related concepts}
        \label{fig:calc_concepts}
    \end{subfigure}
    \hspace{0.03\textwidth}
    \begin{subfigure}{0.3\textwidth}
        \centering
        \includegraphics[width=\linewidth]{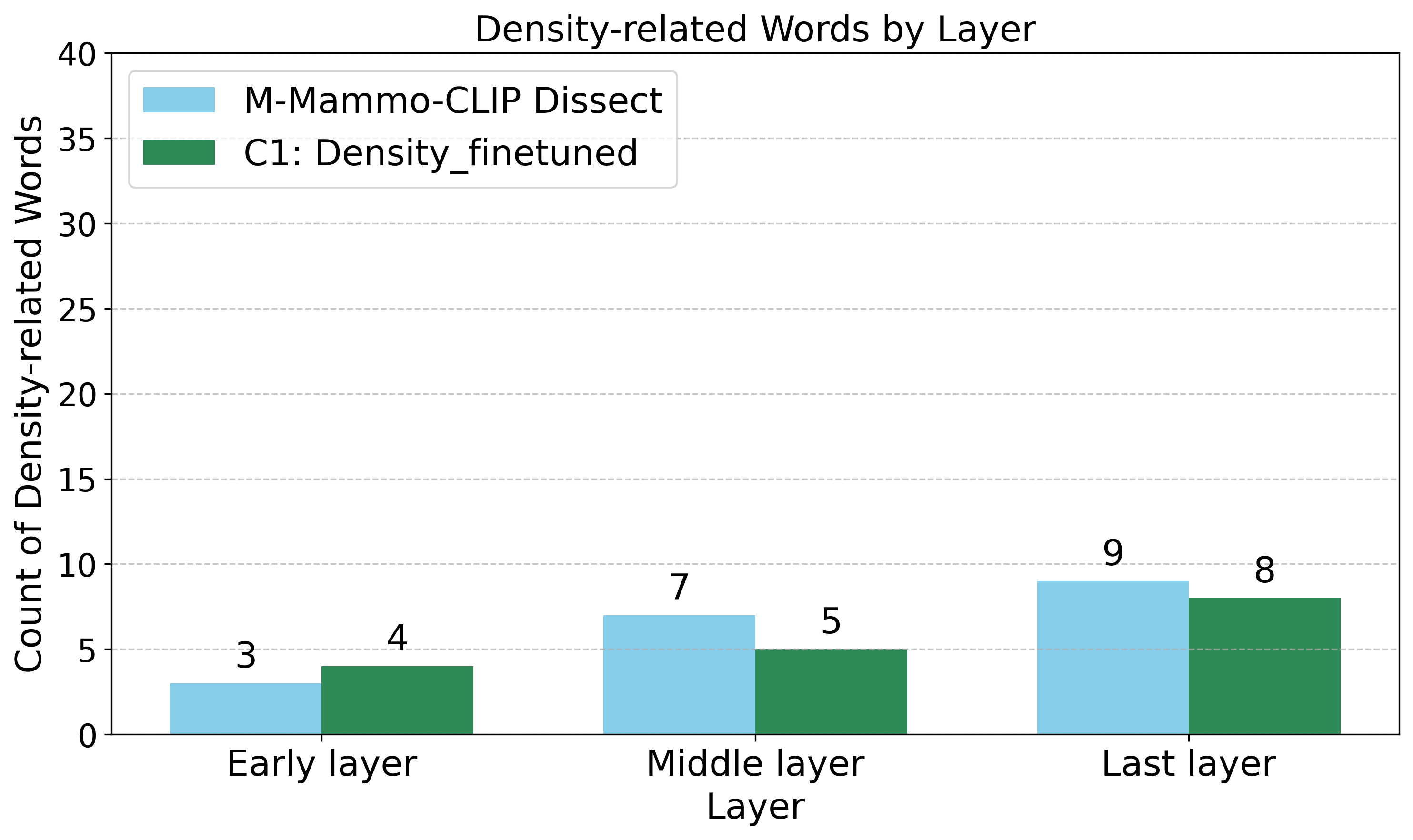}
        \caption{Density related concepts}
        \label{fig:density_concepts}
    \end{subfigure}
    \caption{Comparison of the number of fine-tuning task-related concepts learned by mammo pretrained Mammo-CLIP feature extractors and Mammo-CLIP image classifiers fine-tuned for three different tasks (binary mass, binary suspicious calcification and BI-RADS density).} 
    \label{fig:task_wise_concept_comp}
\end{figure}

Figure \ref{fig:task_wise_concept_comp} compares the number of task-related concepts—specifically for mass, calcification, and density—captured at three representative layers by M-Mammo-CLIP Dissect and the C-Mammo-CLIP Dissect variants. As before, we report the number of unique concepts exceeding the threshold $\tau$. For the mass and density tasks, $F_{target}$ in both models captured a comparable number of relevant concepts, with C-Mammo-CLIP Dissect capturing slightly fewer density-related concepts in the middle and last layers. Interestingly, for calcification, the finetuned $F_{target}$ in C-Mammo-CLIP Dissect identified more task-relevant concepts across the middle and last layers.

To further investigate, we compared the calcification-related concepts uniquely captured by the calcification-finetuned C-Mammo-CLIP Dissect and by M-Mammo-CLIP Dissect, focusing on the last $F_{target}$ layer. Across both models, all calcification-related concepts fell into either the \emph{Calcifications morphology} or \emph{Suspicious calcifications} subcategories, belonging to the broader \emph{Findings and Characterizations} and \emph{Interpretations} categories, respectively. While both models captured the same number of concepts from the \emph{Suspicious calcifications} subcategory, the calcification-finetuned C-Mammo-CLIP Dissect identified more concepts related to \emph{Calcifications morphology} (seven versus four). The specific \emph{Calcifications morphology} concepts captured are listed in Table \ref{Table:calc}. The larger number of \emph{Calcifications morphology} concepts captured by the finetuned model suggests that its $F_{target}$ may be more specialised for distinguishing between calcifications.

\begin{table}[!h]
\caption{Closer look at selected layers of  M-Mammo-CLIP Dissect and C4, calcification fine-tuned, variant of C-Mammo-CLIP Dissect to analyse which task-related concepts were learnt by both and which were unique to each model. Only concepts meeting the mean similarity threshold are included.}
\resizebox{\textwidth}{!}{%
\begin{tabular}{cccc}
\hline
\textbf{Layer} & \textbf{Unique to M-Mammo-CLIP Dissect}  & \textbf{Unique to C4 variant of C-Mammo-CLIP Dissect}   \\ \hline
Last layer        & \begin{tabular}[c]{@{}c@{}}
suture calcification, \\
amourphous calcification,\\ rod-like
\end{tabular} & \begin{tabular}[c]{@{}c@{}}
calcifications in a mass, \\ dystrophic calcification,\\ skin calcification, \\
diffusely scattered punctate calcification,\\ secretory calcification, \\
rim-like calcification, \\
eggshell calcification\end{tabular} &  \\ \hline
\end{tabular}}
\label{Table:calc}
\end{table}
\begin{figure}[!ht]
    \centering
    \includegraphics[width=0.7\linewidth]{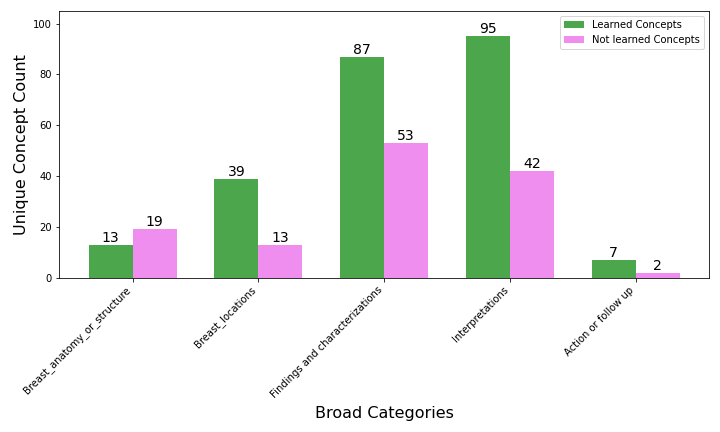}
    \caption{Bar plots showing broad category distribution of number of unique concepts captured versus missed by $F_{target}$ in M-Mammo-CLIP Dissect.}
    \label{fig:learned_ft_bar_both}
\end{figure}

\subsection*{\textbf{RQ3: Which key mammography concepts are learned and which are not picked up by CNNs?}}
In this section, we address our final research question: which mammography concepts are learned—and which are missed—by CNNs trained on mammography data. For this analysis, we focus on $F_{target}$ in M-Mammo-CLIP Dissect and the four variants of C-Mammo-CLIP Dissect, using the VinDR-Mammo test set as probing data.
\begin{figure}[!ht]
    \centering
    \includegraphics[width=0.8\linewidth]{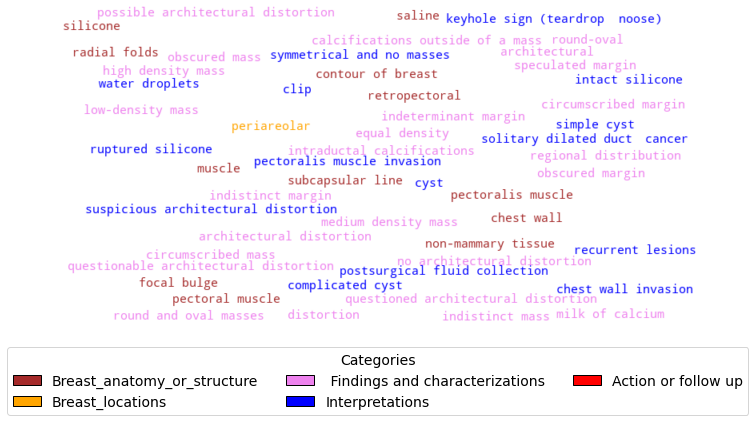}
    \caption{The $54$ unique and distinct mammography concepts not learned by M-Mammo-CLIP Dissect.}
    \label{fig:not_learned_word}
\end{figure}

Overall, $298$ unique concepts were learned and $465$ unique concepts were not learned by M-Mammo-CLIP Dissect, respectively. Among the not learned concepts, $129$
were unique mammography-related concepts, while the remainder belonged to the \emph{Environmental and Natural} categories. In our concept set, some concepts contain overlapping terms, such as \emph{"extremely"} and \emph{"extremely dense,"} or \emph{"amorphous"} and \emph{"amorphous calcification"}. To ensure no overlap between the unique mammography concepts in the learned and not learned sets, we excluded concepts from the not learned set that appeared in a slightly different form in the learned set. This refinement resulted in unique, not learned mammography-related concepts that are distinct from the learned concepts. These 
unique, not learned mammography-related concepts are presented in Figure \ref{fig:not_learned_word}.

We further assess the number of unique mammography concepts captured versus missed by $F_{target}$ in M-Mammo-CLIP Dissect within each of the six broad categories in Figure \ref{fig:learned_ft_bar_both}. The grouping of concepts into broad categories facilitates analysis and helps align comparisons with radiological relevance. Figure \ref{fig:learned_ft_bar_both} shows that, across all broad categories, $F_{target}$ captures more concepts than it misses, indicating that M-Mammo-CLIP Dissect effectively learns most mammography-relevant concepts, with only a small subset remaining uncaptured. We further discuss the implications of the concepts which were not learned in the Discussion section.
\section*{Discussion}
Our findings show that domain-specific training substantially improves alignment with mammography concepts. As shown in Figure \ref{fig:global_new_means_ft_extract}, the $F_{target}$ CNN trained on mammography data (M-Mammo-CLIP Dissect) captured substantially more mammography-specific concepts than the $F_{target}$ trained on ImageNet (G-Mammo-CLIP Dissect). Higher mean similarity values ($\tau$) between neurons and concepts reflect stronger alignment with clinically relevant features. We found M-Mammo-CLIP Dissect captures both more mammography-related and non-mammography-related concepts than G-Mammo-CLIP Dissect in Figure \ref{fig:global_side_by_side_binary_comp}. This suggests that training on mammography data enhances the $F_{target}$ model’s ability to capture mammography-relevant concepts while still learning general ones. Importantly, the consistent top concept categories captured by M-Mammo-CLIP Dissect were \emph{Findings and Characterizations} and \emph{Interpretations}, which mirror radiologists’ workflows. Whereas G-Mammo-CLIP Dissect lacked such focus and switched between capturing non-mammography and mammography concept categories. This is showcased in Figure \ref{fig:global_top3_broad_ft_comp}. In Figure \ref{fig:comparison_last_wordcloud}, we show that G-Mammo-CLIP Dissect focused on more non-mammography concepts than M-Mammo-CLIP Dissect, like \emph{Shih-Tzu} and \emph{goblet}, underscoring that the non-mammography-specific $F_{target}$ may not have effectively learned to represent mammography-specific concepts, likely due to the lack of domain-specific training. Yet some concepts in M-Mammo-CLIP Dissect—such as \emph{stable calcification} or \emph{asymmetrically enlarged lymph node}—may be keyword-driven rather than reflecting clinically meaningful representations, as these require multi-view or temporal analysis, which we have not investigated.

Interestingly, Mammo-CLIP Dissect also reveals individual neurons specialised to distinct mammography features or artefacts. In Figures \ref{fig:implant451} and \ref{fig:calc242}, we observe that the top $5$ concepts captured by the neurons visualised all contain a common keyword. The keywords, in Figures \ref{fig:implant451} and \ref{fig:calc242},  are \emph{implant} and \emph{calcification,} respectively. 

Notably, three of the top five concepts (\emph{benign skin calcification}, \emph{cluster calcification} and \emph{benign calcification}) in Figure \ref{fig:calc242} are indeed represented in the top five images as determined by consultation with a radiologist. However, of the remaining two concepts, \emph{stable calcification} cannot be determined with only one image from a single time-point, and \emph{secretory calcification} does not appear on the top images. This could indicate that, despite concepts comprising several words, the model may sometimes place a greater emphasis on a keyword (i.e., \emph {calcification}) and not truly capture/distinguish the more granular clinical findings/details. This might also occur in the case of the concept \emph{peri-implant fluid} in Figure \ref{fig:implant451}. \emph{peri-implant fluid} is not a concept which can be observed from mammograms, but it is likely captured by the neuron due to the presence of the word \emph{implant} within the concept. 

We also examined how fine-tuning for mammography classification tasks affects the concepts captured by Mammo-CLIP, comparing C-Mammo-CLIP Dissect (four task-specific fine-tuned models) to the base feature extractor in M-Mammo-CLIP Dissect. All fine-tuned models showed higher overall similarity scores $\tau$ than M-Mammo-CLIP Dissect (Figure \ref{fig:new_means_downstream}), indicating that fine-tuning increases alignment between neurons and mammography concepts and increases task specialisation. When comparing the number of unique concepts across key diagnostic categories, Figure \ref{fig:task_wise_concept_comp}, we found little change in mass-related concepts, a slight decrease in density-related concepts, but a clear increase in calcification-related concepts. In particular, the model fine-tuned for suspicious calcifications (C4) captured more benign calcification concepts—such as \emph{dystrophic calcification}, \emph{rim-like calcification}, \emph{eggshell calcification}, \emph{skin calcification} and \emph{secretory calcification}—than M-Mammo-CLIP Dissect, which instead retained the malignant amorphous calcification concept (Table \ref{Table:calc}). 

This pattern suggests that fine-tuning may enhance recognition of certain benign findings while narrowing coverage of other clinically important ones. Consulting radiologists note that benign calcifications can sometimes appear larger and denser than malignant calcifications, making them visually identifiable. This could be a possible reason for the stronger focus of the fine-tuned $F_{target}$ model on the benign findings. But at the same time, distinguishing benign from malignant calcifications remains inherently difficult—even for experienced radiologists— so the observed shift may actually reflect a focus on the keyword \emph{calcification} rather than on fine-grained pathological details.

Understanding the differences in learned and not learned concepts in the $F_{target}$ in M-Mammo-CLIP Dissect is important for identifying the clinically relevant aspects that such models capture, as well as the gaps that may limit their accuracy and trustworthiness. Our analysis in Figure \ref{fig:learned_ft_bar_both} demonstrates that the mammography-trained M-Mammo-CLIP Dissect does capture most concepts relevant to mammography, across all five broad mammography categories present in $\mathcal{C}$. When focusing on the mammography concepts missed by M-Mammo-CLIP Dissect, we find $54$ distinct concepts which are highlighted in Figure \ref{fig:not_learned_word}. We find that these missed concepts include:
\begin{itemize}
    \item Less visually distinct and well-represented findings such as \emph{architectural distortion}, which, while clinically relevant, can be less prominent on mammograms compared to other findings such as masses and calcifications. This concept also may not be well-represented in the training data used. In total, we identified six concepts containing \emph{architectural} or \emph{distortion} which were not captured. We hypothesise that the $F_{target}$ in M-Mammo-CLIP Dissect perhaps fails to grasp this concept due to its under-representation and nuances.
    \item Modality-inappropriate concepts which are linked to ultrasound and MRI and not mammography. This comprises $24$ concepts of the $54$ missed concepts. Some examples for concepts linked to ultrasound and MRI include: \emph{subcapsular line}, \emph{saline}, \emph{silicone} and \emph{radial folds}. Such concepts are not encountered by the $F_{target}$ in M-Mammo-CLIP Dissect both during training and also within our $D_{probe}$, which consists only of mammograms. This reflects that the model in fact does not largely learn concepts from outside the domain of mammograms. In cases of exceptions to this as we observed with \emph{peri-implant fluid}, maybe linked to individual neurons, gravitating to single keywords—such as \emph{implant} —even when the full concept required broader contextual information, suggesting an area for future refinement.
\end{itemize}

In terms of limitations, we note that our concept set $\mathcal{C}$, underpinning our analysis, reflects the challenges in defining clinically valid mammography concepts. Overlaps (e.g., \emph{densities} vs. \emph{density}), inclusion of concepts requiring multi-view or temporal context (e.g. \emph{asymmetry,} \emph{stable calcification}), and modality-specific terms from ultrasound or MRI (e.g., \emph{peri-implant fluid,} \emph{saline}) may confound model evaluation. Despite these limitations, the framework provides valuable insight into the interaction between neurons and concepts.

\section*{Conclusion}
We propose Mammo-CLIP Dissect, a concept-based explainability framework which enables analysis of concepts learned by DL vision models when processing mammograms. This approach extends CLIP-Dissect to the clinical domain and directly addresses the gap between pixel-level saliency and the concept-level reasoning radiologists use. We leveraged Mammo-CLIP Dissect to enable an investigation of concepts learned by domain-specific CNNs trained with mammography data versus general CNNs trained on non-mammography data, how fine-tuning affects these learned concepts, and where key diagnostic concepts remain missing.

Together, our findings demonstrate that concept-based explainability using Mammo-CLIP Dissect offers insights into how CNNs capture mammography-specific knowledge. By comparing models trained on different data sources and fine-tuning regimes, we show how domain-specific training and task-specific adaptation shape concept learning and reveal where clinically important concepts remain underrepresented. This approach moves beyond pixel-level interpretability toward concept-level understanding, bringing DL systems closer to radiologists’ reasoning processes. Future work should focus on refining concept sets to reduce ambiguity, addressing subtle underrepresented findings, and developing fine-tuning strategies that preserve general conceptual breadth while enhancing task-relevant features—ultimately improving interpretability, trust, and clinical adoption of AI in breast cancer screening.

\bibliography{ref}


\begin{figure}[!ht]
    \centering
    \begin{subfigure}[t]{0.5\linewidth} 
        \centering
        \includegraphics[width=0.9\linewidth]{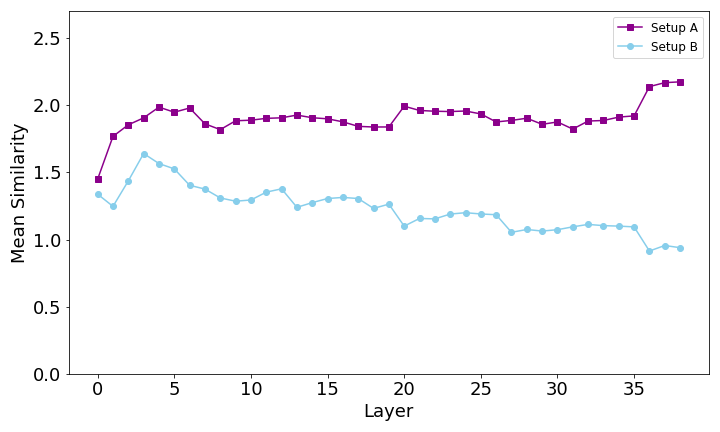}
        \caption{Evolution of mean layer similarities.}
        \label{fig:imgnet_global_new_means_ft_extract}
    \end{subfigure}
    \vskip 1em 
    \begin{subfigure}[t]{0.5\linewidth} 
        \centering
        \includegraphics[width=\linewidth]{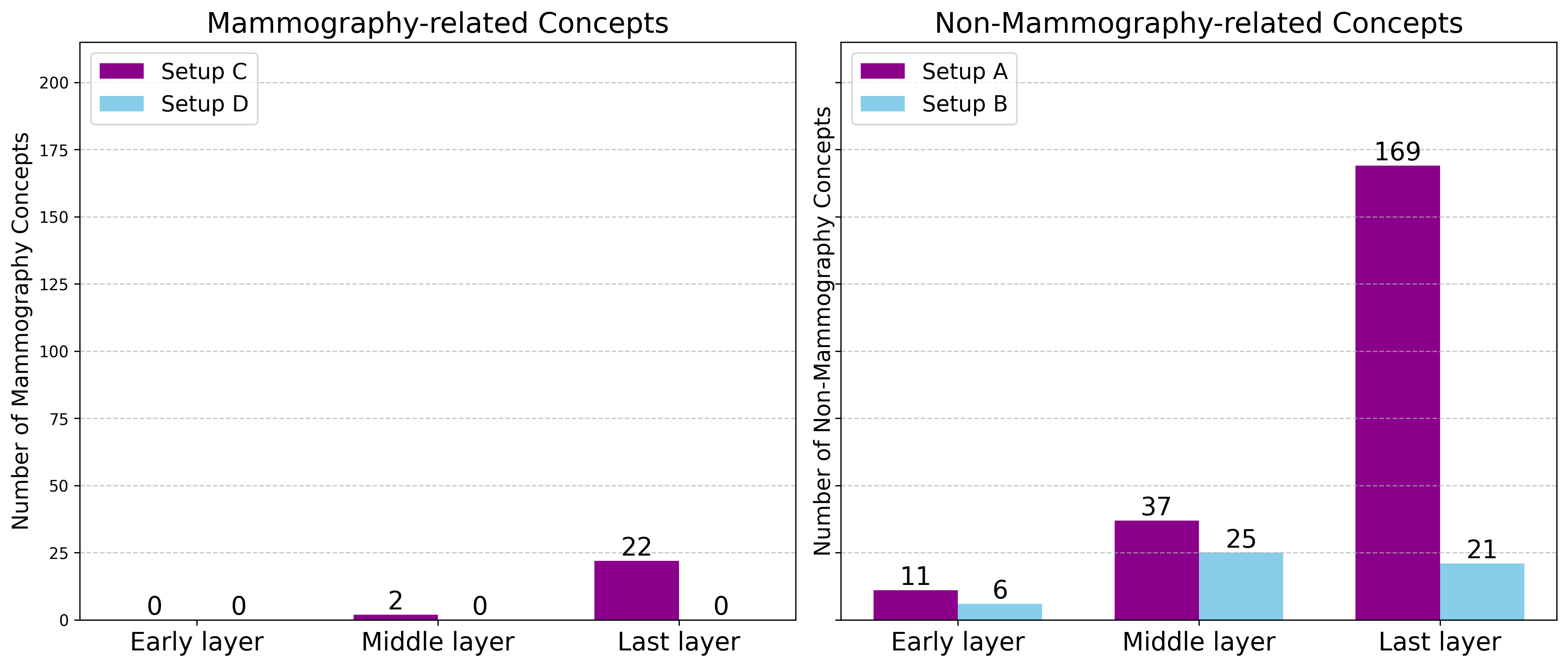}
        \caption{Comparison of occurrences of unique mammography and non-mammography-related concepts.}
        \label{fig:imgnet_global_side_by_side_binary_comp}
    \end{subfigure}
    \hfill
    \begin{subfigure}[t]{0.48\linewidth} 
        \centering
        \includegraphics[width=\linewidth]{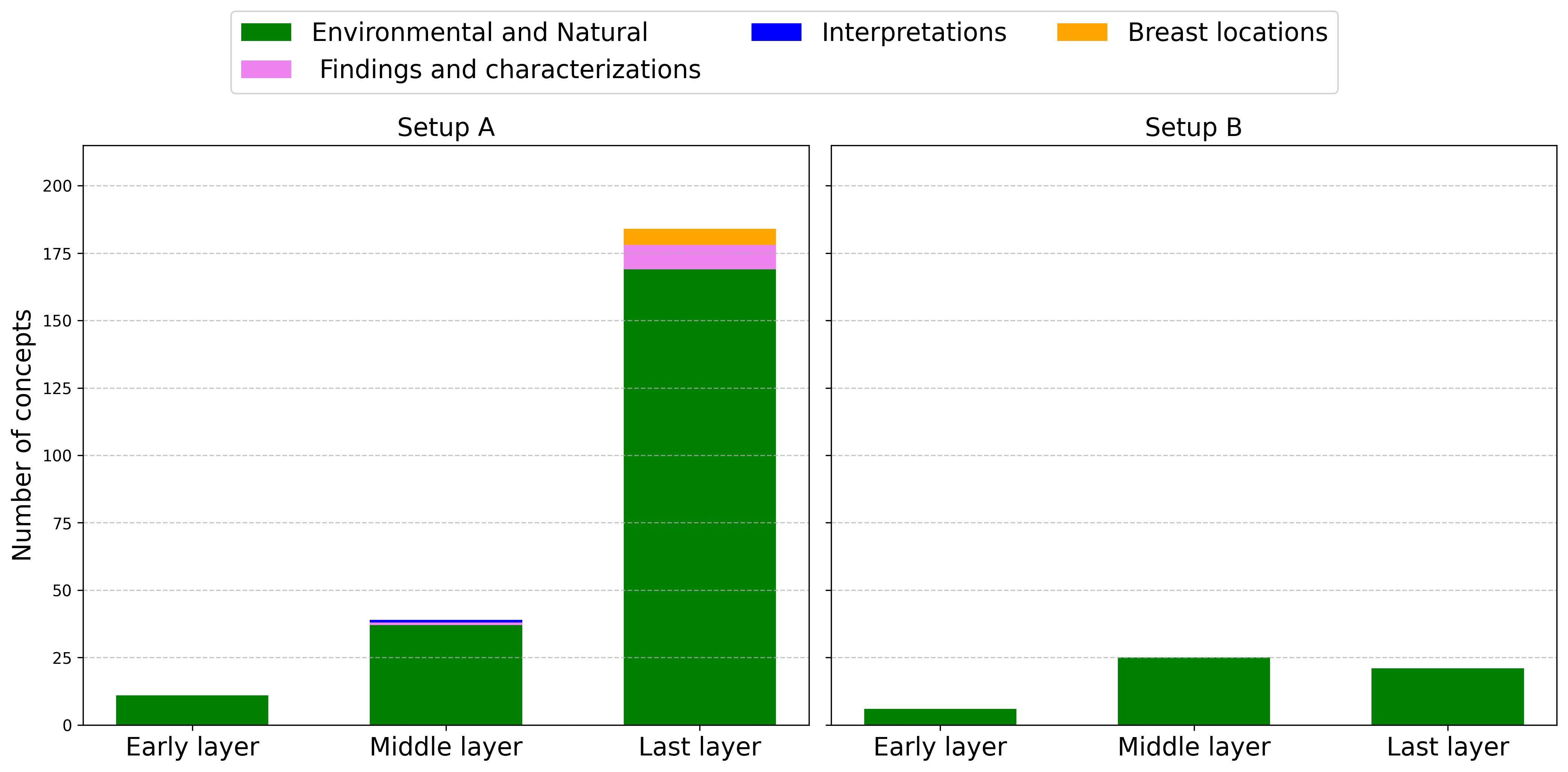}
        \caption{The top three broad concept categories learned by the two models.}
        \label{fig:imgnet_global_top3_broad_ft_comp}
    \end{subfigure}
    \caption{Comparison of Setup A (pretrained on ImageNet) and Setup B (pretrained on mammography) with ImageNet as the probe. Mean layer-specific similarity thresholds have been applied in these plots. (a) Evolution of mean layer similarities, used to determine threshold, $\tau$, across Setup A and Setup B. (b) Occurrences of mammography and non-mammography-related concepts across layers of Setup A and Setup B. (c) Stacked bar plots visualising the top three broad concept categories learned by Setup A on the left and Setup B on the right.}
    \label{fig:app1}
\end{figure}
\section*{Appendix I}
Mammo-CLIP Dissect is a versatile framework that enables flexible choices of the VLM ($F_{dissector}$), the DL vision model under analysis ($F_{target}$), and the probe dataset ($D_{probe}$). In the main text, our analyses focused on configurations where Mammo-CLIP acted as the $F_{dissector}$ and mammography datasets (VinDR-Mammo or EMBED) served as $D_{probe}$. To demonstrate the broader applicability of our approach, we present four alternative experimental setups in which these components are varied. These setups, summarised in Table~\ref{tab:app_setups}, highlight how Mammo-CLIP Dissect can be extended beyond mammography-specific configurations for concept-based analysis.
\begin{table}[h!]
\centering
\resizebox{\columnwidth}{!}{
\begin{tabular}{|c|c|c|>{\centering\arraybackslash}p{8cm}|c|}
\hline
\textbf{Setup name} & \textbf{\( F_{target} \)} & \textbf{\( F_{dissector} \)} & \textbf{Purpose} & \textbf{D\_probe} \\ \hline
Setup A & \( F_{general} \) & CLIP &Examine how a general CLIP dissector combined with an ImageNet-pretrained CNN captures concepts from a natural-image probe. & ImageNet \\ \hline
Setup B & \( F_{mammo} \) & CLIP & Assess how a mammography-pretrained CNN behaves when dissected using a general CLIP dissector and a natural-image probe. & ImageNet \\ \hline
Setup C & \( F_{general} \) & \( F_{general} \) & Compare how a general ImageNet-pretrained CNN model and Mammo-CLIP dissector handle mixed-domain probe data (mammography + natural images). & VinDR-Mammo and ImageNet\\ \hline
Setup D & \( F_{mammo} \) & \( F_{mammo} \) & Examine how a mammography-pretrained CNN model and Mammo-CLIP dissector handle mixed-domain probe data (mammography + natural images). & VinDR-Mammo and ImageNet\\ \hline
\end{tabular}}
\caption{Summary of alternative Mammo-CLIP Dissect configurations we investigated.}
\label{tab:app_setups}
\end{table}
Within these setups, when using ImageNet as the $D_{probe}$, we draw samples from the ImageNette\cite{Imagenette2020} and ImageWoof\cite{Imagenette2020} subsets of ImageNet. Each subset contains 10 classes derived from the full ImageNet dataset. Our $D_{probe}$ comprises 3,000 images from ImageNette and 2,000 images from ImageWoof, with an equal number of images sampled from each of the 10 classes within both subsets.
\begin{figure}[!ht]
    \centering
    \begin{subfigure}[t]{0.5\linewidth} 
        \centering
        \includegraphics[width=0.9\linewidth]{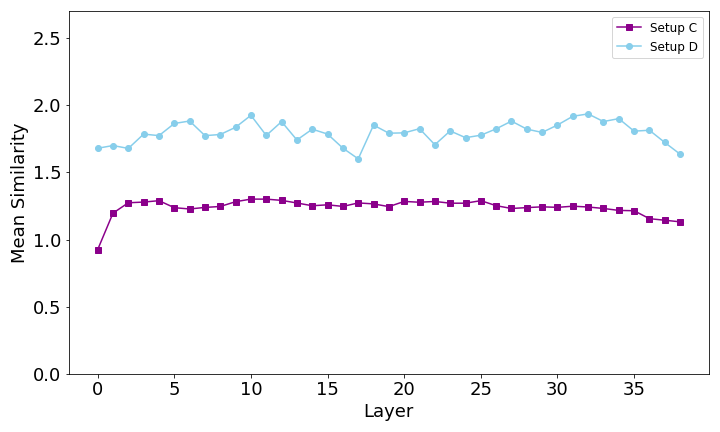}
        \caption{Evolution of mean layer similarities.}
        \label{fig:COMBO_global_new_means_ft_extract}
    \end{subfigure}
    \vskip 1em 
    \begin{subfigure}[t]{0.5\linewidth} 
        \centering
        \includegraphics[width=\linewidth]{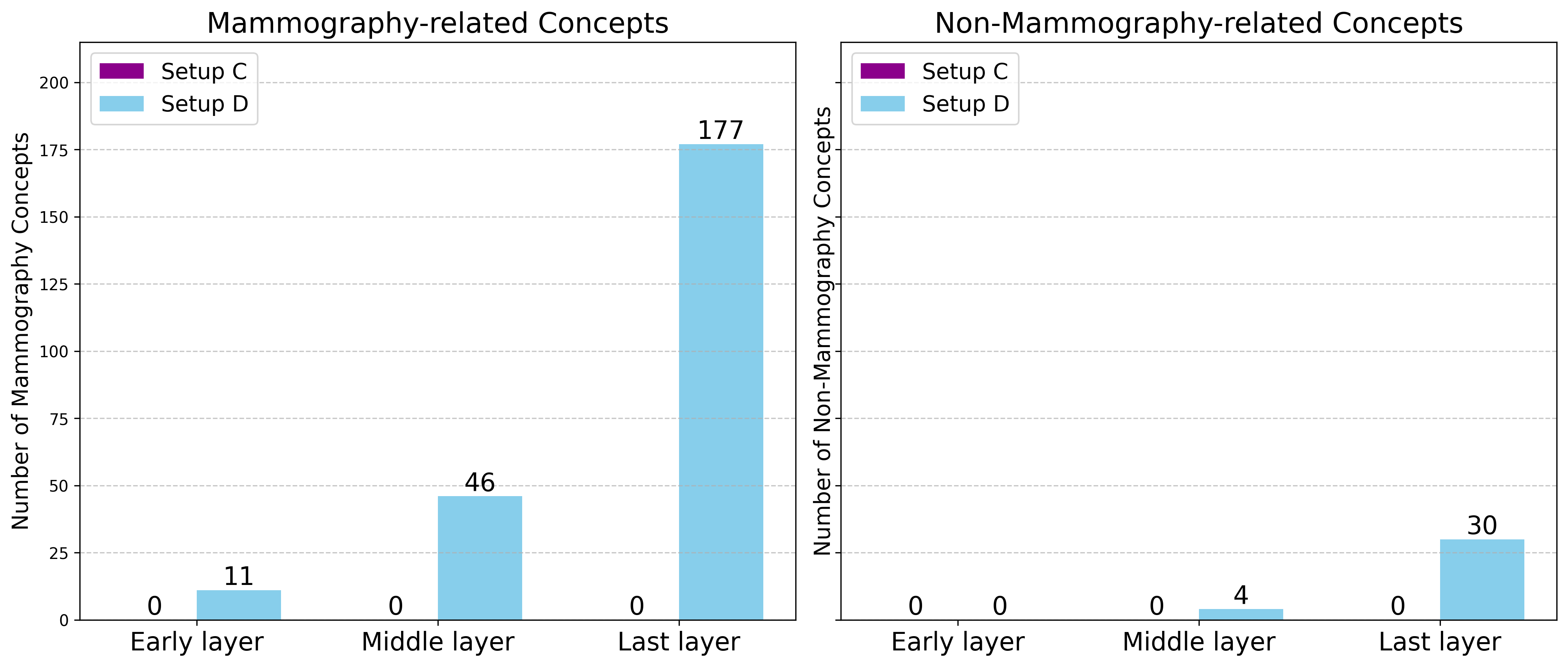}
        \caption{Comparison of occurrences of unique mammography and non-mammography-related concepts.}
        \label{fig:COMBO_global_side_by_side_binary_comp}
    \end{subfigure}
    \hfill
    \begin{subfigure}[t]{0.48\linewidth} 
        \centering
        \includegraphics[width=\linewidth]{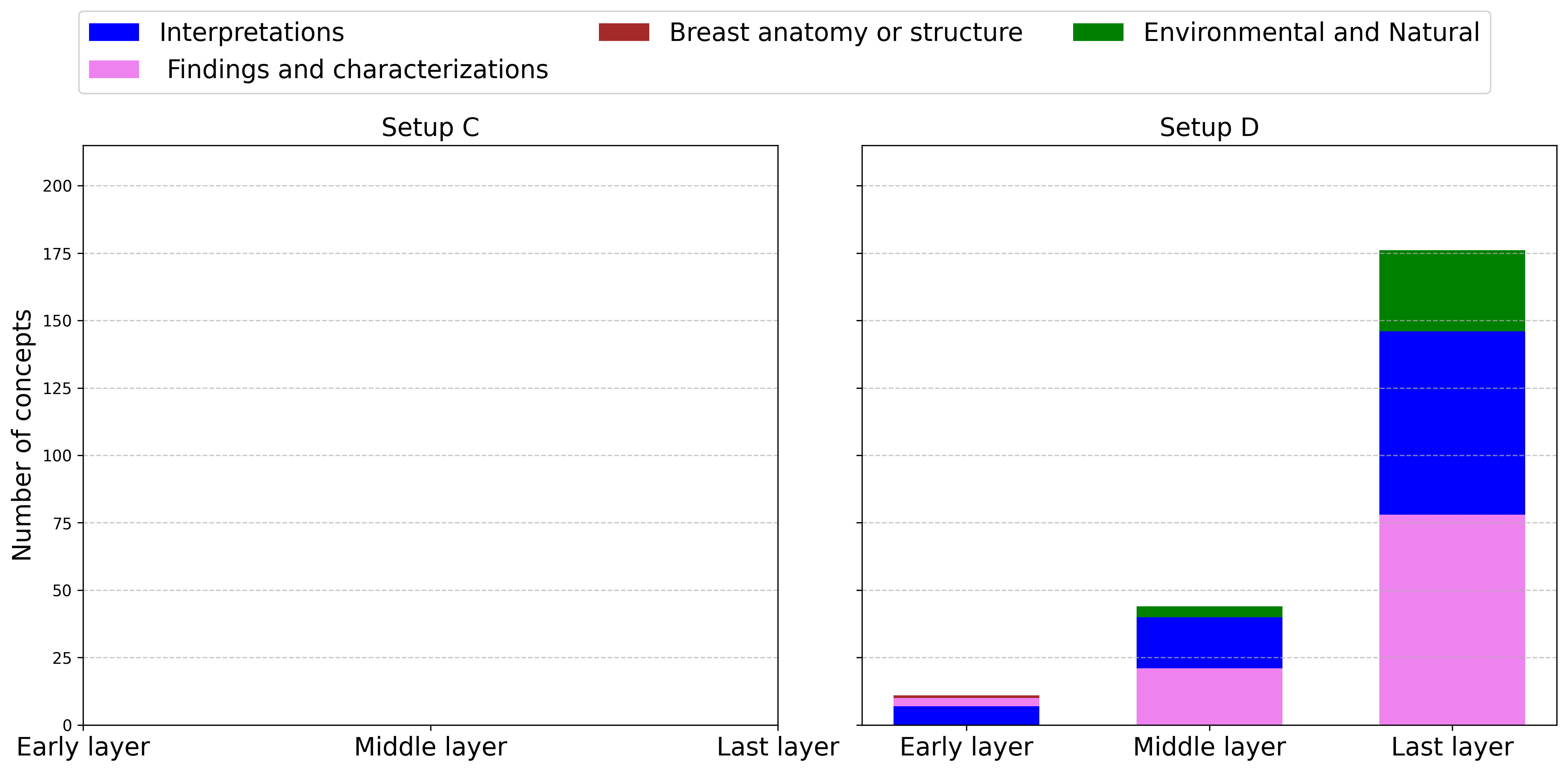}
        \caption{The top three broad concept categories learned by the two models.}
        \label{fig:COMBO_global_top3_broad_ft_comp}
    \end{subfigure}
    \caption{Comparison of Setup C (pretrained on ImageNet) and Setup D (pretrained on mammography) with ImageNet as the probe. Mean layer-specific similarity thresholds have been applied in these plots. (a) Evolution of mean layer similarities, used to determine threshold, $\tau$, across Setup C and Setup D. (b) Occurrences of mammography and non-mammography-related concepts across layers of Setup C and Setup D. (c) Stacked bar plots visualising the top three broad concept categories learned by Setup C on the left and Setup D on the right.}
    \label{fig:app2}
\end{figure}

In Setups A and B, replacing Mammo-CLIP with the general CLIP model as $F_{dissector}$ while using a subset of ImageNet as $D_{probe}$ markedly shifted the distribution of captured concepts toward non-mammography categories. We observe that Setup A with the ImageNet pretrained $F_{target}$ dominates in terms of $\tau$ values (Figure \ref{fig:imgnet_global_new_means_ft_extract}) as well as learning both more of mammography-related and non-mammography-related concepts (Figure \ref{fig:imgnet_global_side_by_side_binary_comp}). Notably, Setup A learns more non-mammography concepts than mammography concepts overall. In Figure \ref{fig:imgnet_global_top3_broad_ft_comp}, we can see that while both setups consistently prioritise the \emph{Environmental and Natural} concepts linked to ImageNet, Setup A, which is trained on ImageNet, captures far more concepts of that category at the last layer.

These findings illustrate how substituting a domain-specific VLM $F_{dissector}$ with a general one amplifies natural-image concepts while perhaps focusing less on the mammography-specific concepts. This highlights the critical role of domain-specific pretraining in steering concept discovery toward clinically meaningful patterns.

Setups C and D mirror the configurations of G-Mammo-CLIP Dissect and M-Mammo-CLIP Dissect, respectively, but replace the purely mammography-based $D_{probe}$ with a mixture of VinDR-Mammo and ImageNet images. This allows us to evaluate how domain-specific versus general pretraining interacts with a mixed probe dataset. Setup D, using an $F_{target}$ trained on mammograms, exhibits higher $\tau$ values than Setup C, suggesting stronger neuron–concept alignment (Figure \ref{fig:COMBO_global_new_means_ft_extract}). Setup D also identifies more mammography-related and non-mammography-related concepts than Setup C (Figure \ref{fig:COMBO_global_side_by_side_binary_comp}) and, when examining the top three concept categories, consistently focuses on \emph{Findings and Characterizations} and \emph{Interpretations}—critical for mammography analysis—while still capturing \emph{Environmental and Natural} concepts linked to ImageNet in the deeper layers (Figure \ref{fig:COMBO_global_top3_broad_ft_comp}). Conversely, Setup C fails to surpass the mean similarity threshold $\tau$ for any concept category, indicating weak overall concept capture. 

These results suggest that mammography-pretrained models retain their domain-specific specialisation even when processing mixed-domain probes, whereas models pretrained solely on general images struggle to form strong concept associations across domains. This may indicate that domain-specific training helps models learn more transferable and robust representations.

\section*{Data availability}
The code implementation and concept set used in this study are made available on GitHub, with the link included in the abstract. The VinDR-Mammo and EMBED probing datasets used in this study are publicly available upon request for access. The EMBED dataset can be accessed from: \url{https://aws.amazon.com/marketplace/pp/prodview-unw4li5rkivs2#links}. Access to the VinDR-Mammo dataset can be obtained from \url{https://physionet.org/content/vindr-mammo/1.0.0/}. In particular, we leveraged the pre-processed VinDR-Mammo images available from \url{https://github.com/batmanlab/Mammo-CLIP}.
\section*{Acknowledgements}
Funded by the Research Council of Norway, Visual Intelligence grant no. 309439, and grant no. 303514.
\section*{Funding}
Funded by the Research Council of Norway, Visual Intelligence grant no. 309439, and grant no. 303514.
\section*{Author contributions statement}

S.A.S., T.D., and K.W. conceived the experiment(s). S.A.S., M.A.M., and T.H. curated the concept set. S.A.S. conducted the experiments. S.A.S., T.D., M.A.M., T.H., A.P., E.W., K.W., R.J., and M.K. analysed the results.  All authors reviewed the manuscript. 

\section*{Additional information}

\textbf{Competing interests}: The authors declare no competing interests.. 


\end{document}